\pdfoutput=1
\documentclass[11pt]{article}

\usepackage{ACL2023}

\usepackage{times}
\usepackage{latexsym}

\usepackage[T1]{fontenc}
\usepackage[utf8]{inputenc}

\usepackage{microtype}

\usepackage{inconsolata}
\usepackage{amsfonts}
\usepackage{amsmath}
\usepackage{graphicx}
\usepackage{subcaption}
\usepackage{amsmath}
\usepackage{amsthm}
\usepackage{booktabs}
\usepackage{algorithm}
\usepackage{algorithmic}
\usepackage{multirow}
\usepackage{multicol}
\usepackage{mathptmx}

\renewrobustcmd{\bfseries}{\fontseries{b}\selectfont}
\renewrobustcmd{\boldmath}{}
\newrobustcmd{\B}{\bfseries}
\title{DM-Align: Leveraging the Power of Natural Language Instructions to Make Changes to Images}

\author{Maria Mihaela Trusca\textsuperscript{1}, Tinne Tuytelaars\textsuperscript{2}, Marie-Francine Moens\textsuperscript{1} \\
\textsuperscript{1}KU Leuven, Department of Computer Science\\
\textsuperscript{2}KU Leuven, Department of Electrical Engineering\\
\texttt{\{mariamihaela.trusca, tinne.tuytelaars, sien.moens\}}@kuleuven.be
}

\begin{document}
% \nolinenumbers
\maketitle

\begin{abstract}

Text-based semantic image editing assumes the manipulation of an image using a natural language instruction. Although recent works are capable of generating creative and qualitative images, the problem is still mostly approached as a black box sensitive to generating unexpected outputs. Therefore, we propose a novel model to enhance the text-based control of an image editor by explicitly reasoning about which parts of the image to alter or preserve. It relies on word alignments between a description of the original source image and the instruction that reflects the needed updates, and the input image. The proposed Diffusion Masking with word Alignments (DM-Align) allows the editing of an image in a transparent and explainable way. It is evaluated on a subset of the Bison dataset and a self-defined dataset dubbed Dream. When comparing to state-of-the-art baselines, quantitative and qualitative results show that DM-Align has superior performance in image editing conditioned on language instructions, well preserves the background of the image and can better cope with long text instructions. 

\end{abstract}

\section{Introduction}

AI-driven image generation was confirmed as a smooth-running option for content creators with high rates of efficiency and also creativity \citep{DBLP:journals/corr/abs-2204-06125} that can be easily adapted to generate consecutive frames for video generation \citep{DBLP:conf/nips/0004ZH022, DBLP:conf/iclr/SingerPH00ZHYAG23}. Text-based guidance has proven to be a natural and effective means of altering visual content in images. Various model architectures have been proposed for text-based image synthesis, ranging from transformers \citep{DBLP:conf/nips/DingYHZZYLZSYT21, DBLP:conf/nips/VaswaniSPUJGKP17} to generative adversarial networks (GANs) \citep{DBLP:conf/nips/GoodfellowPMXWOCB14, DBLP:conf/icml/ReedAYLSL16, DBLP:conf/cvpr/ZhuP0019}, and more recently, diffusion models like DALL·E 2 \citep{DBLP:journals/corr/abs-2204-06125}, Imagen \citep{DBLP:journals/corr/abs-2205-11487}, or Stable Diffusion Models \citep{DBLP:conf/cvpr/RombachBLEO22}. The success of diffusion models, akin to that observed in language models \citep{DBLP:journals/corr/abs-2001-08361}, largely results from their scalability. Factors such as model size, training dataset size, and computational resources contribute significantly to their effectiveness, overshadowing the impact of the model architecture itself. This scalability enables these models to adapt easily to different domains, including unseen concepts \citep{DBLP:conf/icml/RameshPGGVRCS21, DBLP:journals/corr/abs-2205-11487}. Moreover, these models are ready to use without the need for additional training \citep{DBLP:conf/iccv/ChoiKJGY21, DBLP:conf/cvpr/LiQLT20}. 

\begin{figure}[t]
\begin{center}
   \includegraphics[width=\linewidth]{"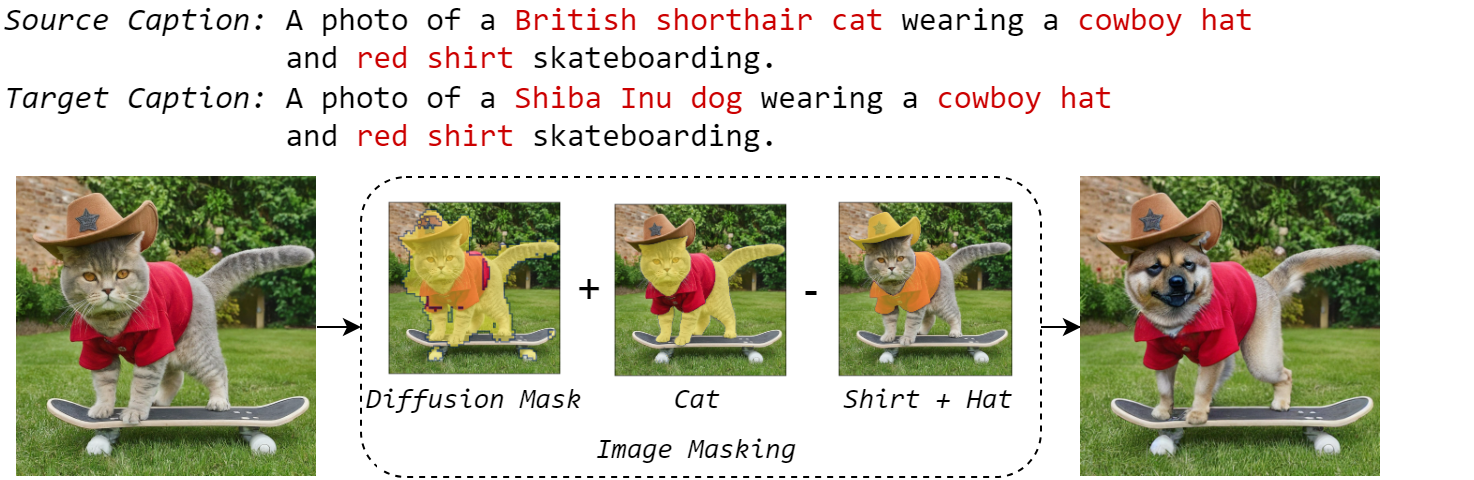"}
\end{center}
   \caption{The proposed image editor utilizes a source caption to describe the initial image and a target text instruction to define the desired edited image. To accomplish this task, we employ the two captions to generate a diffusion mask, refining it further by incorporating regions of words that we want to keep or alter in the image.}
\label{fig:teaser}
\end{figure}

\begin{figure*}[t]
\begin{center}
   \includegraphics[width=\textwidth]{"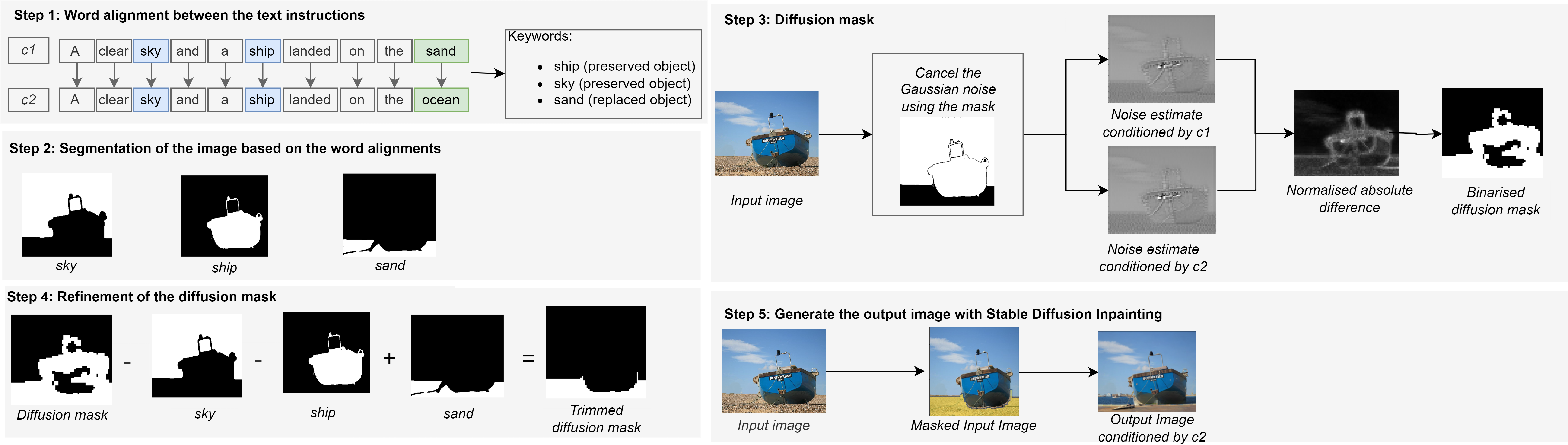"}
\end{center}
   \caption{The implementation of DM-Align. The aim is to update the input image described by the text instruction $c_1$ (``A clear sky and a ship landed on the sand") according to the text instruction $c_2$ (``A clear sky and a ship landed on the ocean").}
\label{fig:overview}
\end{figure*}

While similar to the text-based semantic image generation task in its creation of new visual content, text-guided image editing also relies on additional visual guidance. Consequently, the goal of text-guided image manipulation is to modify the content of a picture according to a given text while keeping the remaining visual content untouched. The remaining visual content is from now on referred to as ``background". As text-to-image generators, text-based image editors work at the frame level and can be further adapted for video editing \citep{DBLP:conf/nips/ZhangLNHGL23}. Text-based semantic image editing typically employs text-based image generation models with user-defined image masks \citep{DBLP:journals/corr/abs-2206-02779, DBLP:conf/cvpr/AvrahamiLF22, DBLP:journals/corr/abs-2212-06909, DBLP:journals/corr/abs-2212-05034}. Each of these masks is an arrangement that differentiates between the image content that is to be changed or preserved. 
However, asking humans to generate masks is cumbersome, so we would like to edit images naturally, relying solely on a textual description of the image and its instruction to change it. Existing models for text-based semantic image editing, which do not require human-drafted image masks, struggle to maintain the background \citep{DBLP:conf/cvpr/BrooksHE23, couairon2022flexit, DBLP:conf/cvpr/KawarZLTCDMI23, DBLP:conf/cvpr/TumanyanGBD23, DBLP:journals/corr/abs-2302-05543}. Preserving the background's consistency is particularly relevant for applications like game development or virtual worlds, where visual continuity across frames is crucial. Finally, the complexity of textual instructions given by their length poses a challenge for semantic image editors. While the existing models can effectively handle short text instructions, they encounter difficulties in manipulating an image using longer and more elaborate ones.

To address the aforementioned limitations, we present a novel approach that employs one-to-one alignments between the words in the text instruction describing the source image and those describing the desired edited image (Figure \ref{fig:teaser}). By leveraging these word alignments, we implement image editing as a series of deletion, insertion, and replacement operations. Through this text-based control mechanism, our proposed model consistently produces high-quality editing results, even with long text instructions, while ensuring the preservation of the background. 

\begin{figure*}[t]
\centering
   {   \includegraphics[width=14cm]
   {"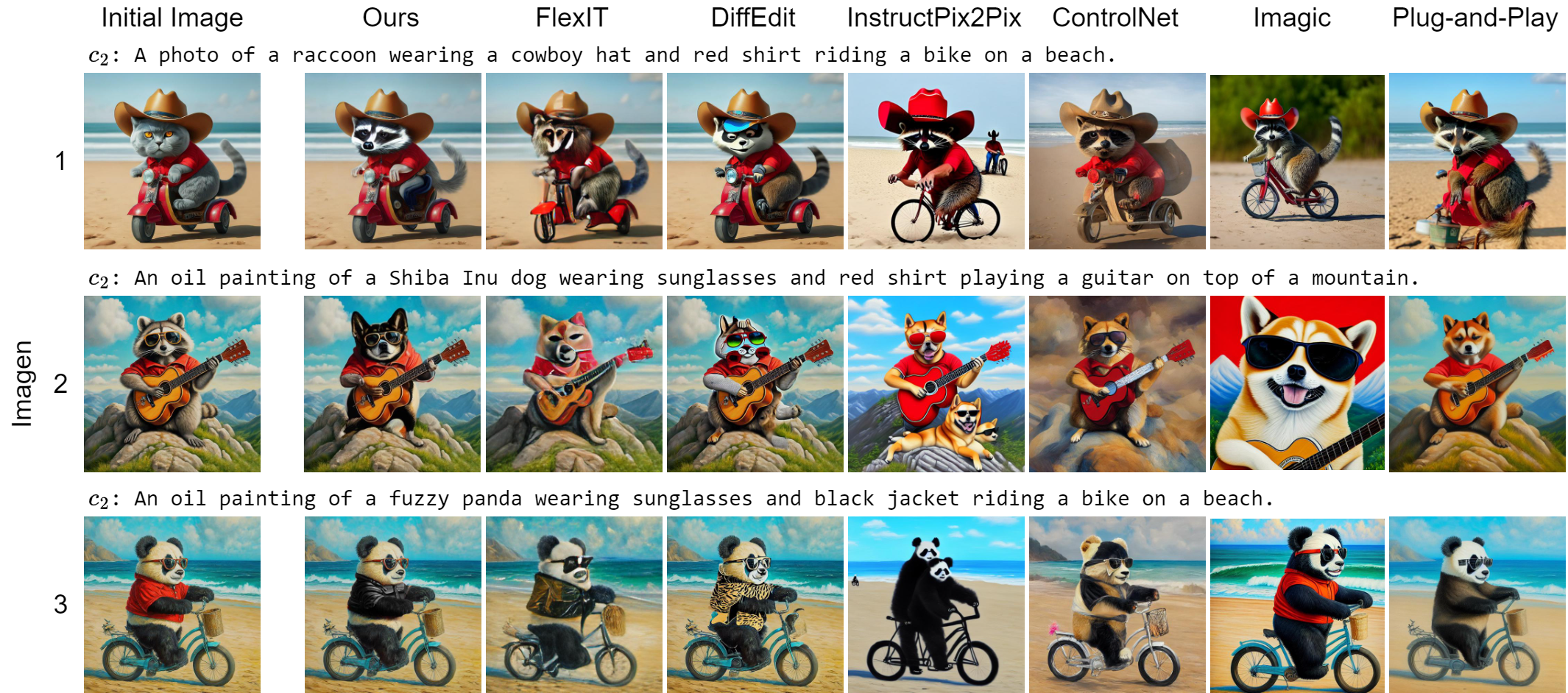"}}
  { \caption{Semantic image editing: Imagen dataset. Source captions: (1) $c_1$. A photo of a British shorthair cat wearing a cowboy hat and red shirt riding a bike on a beach. (2) $c_1$. An oil painting of a raccoon wearing sunglasses and red shirt playing a guitar on top of a mountain. (3) $c_1$. An oil painting of a fuzzy panda wearing sunglasses and red shirt riding a bike on a beach. 
   }}\label{fig:examples_imagen}
\end{figure*}

As presented in Figure \ref{fig:overview}, we align the words of the text that describes the source image and the textual instruction that describes how the image should look after the editing, which allows us to determine the information the user wants to keep, or replace. Then, disjoint regions associated with the preserved or discarded information are detected by segmenting the image. Next, a global, rough mask for inpainting is generated using standard diffusion models. While the diffusion mask allows the insertion of new objects that are larger than the replaced ones, it has the disadvantage of being too rough. Therefore, we further refine it using again the detected disjoint regions. To prove the effectiveness of DM-Align, the masked content is generated using inpainting stable diffusion \citep{DBLP:conf/cvpr/RombachBLEO22}.

Our contributions are summarized as follows:
\begin{enumerate}
  \item Our novel approach reasons with the text caption of the original input image and the text instruction that guides the changes in the image, which is a natural and human-like way of approaching image editing with a high level of explainability.
  \item By differentiating between the image content to be changed from the content to be left unaltered, the proposed DM-Align enhances the text control of semantic image editing.
  \item Compared to recent models for text-based semantic image editing, DM-Align demonstrates superior capability in handling long text instructions and preserving the background of the input image while accurately implementing the specified edits.

\end{enumerate}

\section{Related work}\label{related_works}

Despite the aim of keeping the background as similar as possible to the input image, numerous AI-based semantic image editors insert unwanted alterations in the image. FlexIt \citep{couairon2022flexit} combines the input image and instruction text into a single target point in the CLIP multimodal embedding space and iteratively transforms the input image toward this target point. %In another approach \citep{DBLP:journals/corr/abs-2209-15264}, the image editing is framed as an image translation task, relying on style, and structure losses to guide the training of the model.
\citet{DBLP:journals/corr/abs-2302-05543} introduce ControlNet as a neural network based on two diffusion models, one frozen and one trainable. While the trainable model is optimized to inject the textual conditionality of the semantic editing, the frozen model preserves the weights of the model pre-trained on large image corpora. The output of ControlNet is gathered by summing the outputs of the two diffusion models. To keep the structural information of the input image, \citet{DBLP:conf/cvpr/TumanyanGBD23} define a Plug-and-Play model as a variation of the Latent Diffusion Model. Their method edits an input image using not only textual guidance but also a set of features that separately store spatial information and layout details like the shape of objects. 

While most text-based image editors are training-free, Imagic proposed by \citet{DBLP:conf/cvpr/KawarZLTCDMI23}, assumes fine-tuning of the diffusion model by iteratively running over a text embedding optimized to match the input image and resemble the editing text instruction. Ultimately, the text embedding of the editing instructions and the optimized text embedding are interpolated and utilized as input by the fine-tuned model to generate the final edited image. This idea of fine-tuning a diffusion model is also adopted by \citet{DBLP:conf/cvpr/BrooksHE23} to define InstructPix2Pix as a model that approaches text-based image editing as a supervised task. Due to the scarcity of data, a methodology relying on Prompt-to-Prompt \citep{DBLP:journals/corr/abs-2208-01626} is proposed for generating pairs of images before and after the update. During inference, the fine-tuned stable diffusion model can seamlessly edit images using an input image and a text instruction.

The above approaches lack an explicit delineation of the image content to be altered. Closer to our work is the Prompt-to-Prompt model \citep{DBLP:journals/corr/abs-2208-01626} which connects the text prompt with different image regions using cross-attention maps. The image editing is then performed in the latent representations responsible for the generation of the images. In contrast, our work focuses on the detection and delineation of the content to be altered in the image and is guided by the difference in textual instructions. Additionally, we edit images using real pictures and not latent representations artificially generated by a source prompt. 

To overcome the problem of unwanted alterations in the image, DiffEdit \citep{DBLP:journals/corr/abs-2210-11427} computes an image mask as the difference between the denoised outputs using the textual instruction that describes the source image and the instruction that describes the desired edits. However, without an explicit alignment between the two text instructions and the input image, DiffEdit has little control over the regions to be replaced or preserved. While DiffEdit internally creates the editing mask, models like SmartBrush \citep{DBLP:journals/corr/abs-2212-05034}, Imagen Editor \citep{DBLP:journals/corr/abs-2212-06909}, Blended Diffusion \citep{DBLP:conf/cvpr/AvrahamiLF22} or Blended Latent Diffusion \citep{DBLP:journals/corr/abs-2206-02779} directly edit images using hand-crafted user-defined masks.

Due to a rough text-based control, the above models often struggle with preserving background details and are overly sensitive to the length of text instructions. Different from the current models, our DM-Align model does not treat the recognition of the visual content that requires preservation or substitution as a black box. By explicitly capturing the semantic differences between the natural language instructions, DM-Align provides comprehensive control over image editing. This novel approach results in superior preservation of unaltered image content and more effective processing of long text instructions. Except for the models that require additional input masks, all the above-mentioned text-based image editors are used as baselines for our evaluation. 

\section{Proposed model}\label{proposed_model}

In this section, we present our solution for semantic image editing. We define the task and then describe the main steps of the proposed model, which consist of: 1) Detecting the content that needs to be updated or kept relying on the alignment of words of the text that describes the source image and the textual instruction that describes how the image should look after the editing; 2) The segmentation of the image content to be updated or kept by cross-modal grounding; 3) The computation of a global diffusion mask that assures the coherence of the updated image; 4) The refinement of the global diffusion mask with the segmented image content that will be updated or kept; and 5) The inpainting of the mask with the help of a diffusion model. As demonstrated by our experiments, the proposed DM-Align can successfully replace, delete, or insert objects in the input image according to the text instructions. Our method mainly focuses on the nouns of the text instructions and their modifiers. Consequently, DM-Align does not implement action changes and the resulting changes in the position or posture of objects in the input image, which we leave for future work.

\subsection{Task Definition}\label{sec:task_definition}

DM-Align aims to alter a picture described by a source text description or instruction $c_1$ using a target text instruction $c_2$. Considering this definition, the purpose is to adjust only the updated content mentioned in the text instruction $c_2$ and leave the remaining part of the image unchanged. Based on this, we argue the need for a robust masking system that clearly distinguishes between unaltered image regions, which we call ``background", and the regions that require adjustments. 

\subsection{Word alignment between the text instructions}\label{sec:mask_refinament}

The alignment represents the first step of the DM-Align model proposed to enhance the text-based control for semantic image editing (Figure \ref{fig:overview}). Given the two text instructions $c_1$ and $c_2$, our assumption is that the shared words should indicate unaltered regions, while the substituted words should point to the regions that require manipulations. Implicitly, the most relevant words for this analysis are nouns due to their quality of representing objects in the picture. The words are syntactically classified using the Stanford part-of-speech tagger \citep{DBLP:conf/naacl/ToutanovaKMS03}.

We extend the region to be edited by including the regions of the shared words with different word modifiers\footnote{A modifier is a word or phrase that offers information about another word mentioned in the same sentence. To keep the editing process simple, in the current work we use only word modifiers represented by adjectives.} in the two text instructions. As a result, the properties of the already existing objects in the picture can be updated. On the contrary, if the aligned nouns have identical modifiers (or no modifiers) in both instructions, their regions in the image should be unaltered. In addition, we also consider the regions of the unaligned nouns mentioned in the source text instruction (deleted nouns) as unaltered regions. Keeping the regions of the deleted nouns is important because we assume that in the target instruction, a user only mentions the desired changes in the image, omitting irrelevant content \citep{hurley2014concise}. Editing the regions of the deleted nouns reduces the similarity w.r.t the source image and increases the level of randomness in the target image since we generate new visual content that is irrelevant to both the source image and the target caption (Figure \ref{fig:delete}).

\begin{figure}[h]
\begin{center}
   \includegraphics[width=7cm]{"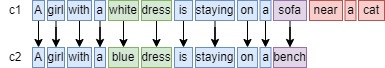"}
\end{center}
   \caption{Word alignment example. Blue: identical words, Purple: substituted words, Green: nouns with different modifiers, Red: nouns mentioned only in the source caption $c_1$.}
\label{fig:word_alignment}
\end{figure}

Considering the example presented in Figure \ref{fig:word_alignment}, the diffusion mask is adjusted to include the regions assigned to the sofa and dress. While the sofa is substituted with a bench, the dress has different modifiers in the captions. On the other hand, the regions of nouns ``girl" and ``cat" are eliminated from the diffusion mask. The girl is mentioned in both captions, while the cat is irrelevant to the user according to the caption $c_2$ and is incorporated in the background.

The detection of word alignments between the two text instructions is realized with a neural semi-Markov CRF model \citep{DBLP:conf/acl/LanJX20}. The model is trained to optimize the word span alignments, where the maximum length of spans is equal to $D$ words (in our case $D$ = 3). The obtained word span alignments will then further be refined into word alignments.
%To convert the word span alignment to word alignments, all words of a span in the source text are connected to all words of a span in the target text. In the end, the word alignments are represented by a set of pairs $(i-j)$, where $i$ is a word of the instruction $c_1$, and $j$ is a word of the instruction $c_2$.

\begin{figure*}[t]
\centering
   {   \includegraphics[width=14cm]
   {"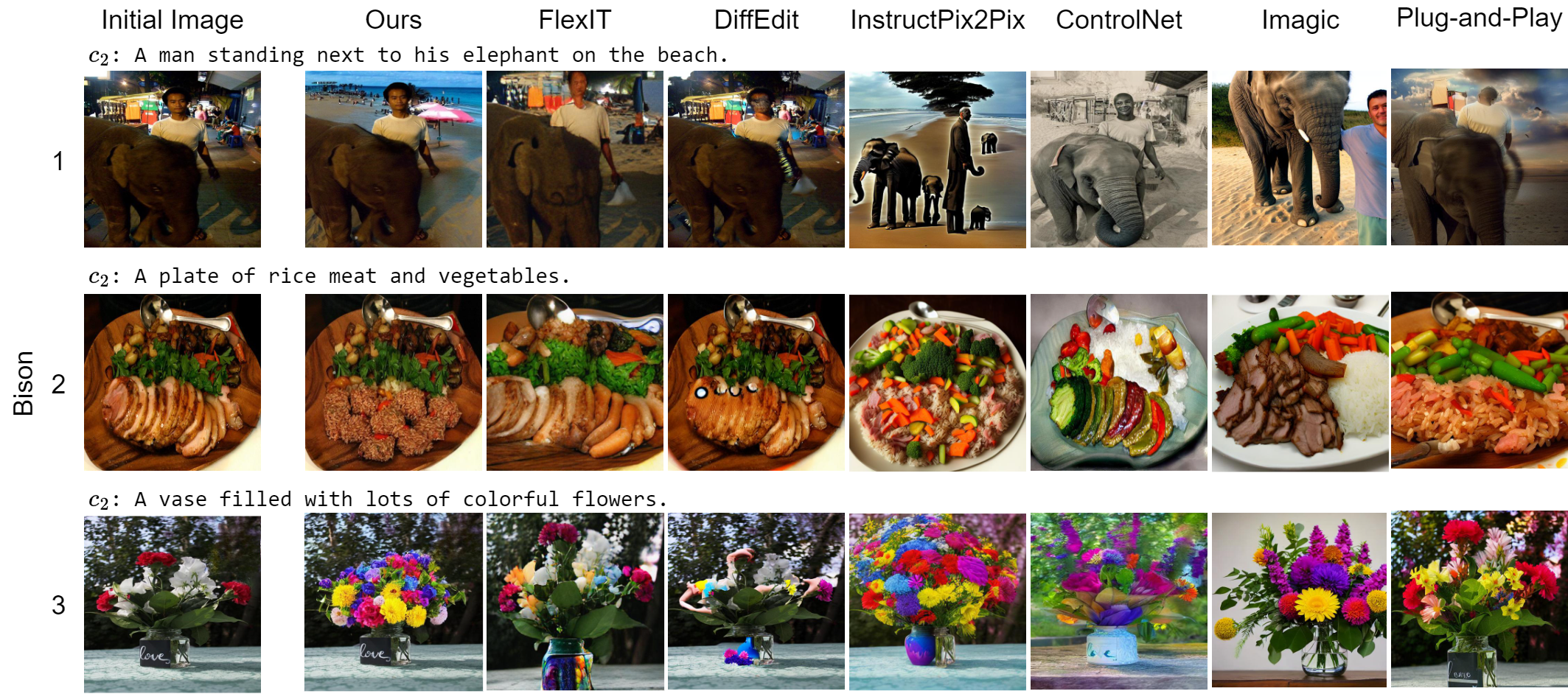"}}
  { \caption{Semantic image editing: Bison dataset. Source captions: (1) $c_1$. A man standing next to a baby elephant in the city. (2) $c_1$. A wooden plate topped with sliced meat and vegetables. (3) $c_1$. A vase filled with red and white flowers.
   }}
   
\label{fig:examples_bison_data}
\end{figure*}

The neural semi-Markov CRF model is optimized to increase the similarity between the aligned source and target word span representations, which are each computed with a pretrained SpanBERT model \citep{joshi-etal-2020-spanbert}. The component that optimizes the similarity between these representations is implemented as a feed-forward neural network with Parametric ReLU \citep{he2015delving}. To avoid alignments that are far apart in the source and target instructions, another component controls the Markov transitions between adjacent alignment labels. To achieve this, it is trained to reduce the distance between the beginning index of the current target span and the end index of the target span aligned to the former source span. Finally, a Hamming distance is used to minimize the distance between the predicted alignment and the gold alignment. The outputs of the above components are fused in a final function $\psi(a|s,t)$ that computes the score of an alignment $a$ given a source text $s$ and target text $t$. The conditional probability of span alignment $a$ is then computed as:  
%Therefore, the model learns to encourage short-distance word alignments. 
%Given the instructions $c_1$ and $c_2$ as the source and target sentences in the neural semi-Markov CRF model, the predicted word alignments\footnote{The word alignments are represented by a set of pairs $(i-j)$, where $i$ is a word of the instruction $c_1$, and $j$ is a word of the instruction $c_2$.} $a$ ($a \in \mathcal{A}$) and the ground truth alignment $a^*$, the model is trained to increase the conditional probability 
\begin{equation}
p(a | s,t) = \frac{e^{\psi(a|s,t)}}{\sum_{a' \in \mathcal{A}}{{e^{\psi(a'|s,t)}}}} %{}/{}
\end{equation}

where the set $\mathcal{A}$ denotes all possible span alignments between source text $s$ and target text $t$. 
The model is trained by minimizing the negative log-likelihood of the gold alignment $a^*$ from both directions, that is, source to target $s2t$ and target to source $t2s$ : 

\begin{equation}
    \sum_{s,t,a^*}{-\mbox{log} \: {p(a^*_{s2t}|s,t)} - \mbox{log} \: {p(a^*_{t2s}|t,s)}}
\end{equation}
The neural semi-Markov CRF model is trained on the MultiMWA-MTRef monolingual dataset, a subset of the MTReference dataset \citep{yao2014feature}. Considering the trained model, we predict the word alignments as follows. Given two text instructions $c1$ and $c2$, the model predicts two sets of span alignments $a$: $a_{s2t}$ aligning $c1$ to $c2$;
%with $c1$ as the source text and $c2$ as the target text;
and $a_{t2s}$ aligning $c2$ to $c1$ 
%with $c2$ as the  text and $c1$ as the source text. 
The final word alignment is computed by merging these two span alignments. Let $i$ be a word of the source text and $j$ be a word of the target text, if alignment $a_{s2t}$ indicates the connection $i->j$ and alignment $a_{t2s}$ indicates the connection $j->i$, then the words $i$ and $j$ become aligned. % o convert the word span alignment to word alignments, all words of a span in the source text are connected to all words of a span in the target text. 
In the end, the word alignments are represented by a set of pairs $(i-j)$, where $i$ is a word of the instruction $c_1$, and $j$ is a word of the instruction $c_2$.
%In this case, i is a word of the source text and j is a word of the target text.
%from the source instruction $c1$ to the target instruction $c_2$ with maximum probability.

%defined 
%based on:
%\begin{equation}
%\begin{array}{l}
%{\psi(a|c_1,c_2)} = \sum_i{\nu({c_1}_i,{c_2}_i)+\tau({c_2}_{i-1}^m,{c_2}_i^1)}+ \\dist(a,a^*)
%\end{array}
%\end{equation}

%\noindent where ${c_1}_i$ and ${c_2}_i$ are the $i-th$ spans of the instructions $c_1$, and $c_2$. ${c_2}_{i-1}^m$ represents the last word at the position $m$ in the previous word span of the target instruction ${c_2}_{i-1}$. ${c_2}_i^1$ indicates the first word of the current word span of the target instruction ${c_2}_i$. The function $dist$ represents the Hamming distance and increases the matching between the real and predicted alignments. 

\subsection{Segmentation of the image based on the word alignments}\label{segmentation_image_word_align}

The aim is to identify the regions in the image that require changes or conservation (second step in Figure \ref{fig:overview}). Based on the above word alignments, we select the nouns whose regions will be edited (non-identical aligned nouns or aligned nouns with different modifiers in the two text instructions) and the nouns whose regions will stay unaltered (nouns of the source text instruction not shared with the target text instruction, identical aligned nouns). Once these nouns are selected we use Grounded-SAM \citep{grounded_sam} to detect their corresponding image regions. Its benefit is the ``open-set object detection" achieved by the object detector Grounding DINO \citep{DBLP:journals/corr/abs-2303-05499} which allows the recognition of each object in an image that is mentioned in the language instruction. Given a noun, Grounding DINO detects its bounding box in the image, and SAM \citep{DBLP:journals/corr/abs-2304-02643} determines the region of the object inside the bounding box. The selected regions will be used to locally refine the diffusion masks discussed in the next section.  

\begin{table*}[t]
\small
\caption{Image-level evaluation for Dream, Bison and Imagen datasets (mean and variance). Compared with the baselines, DM-Align achieves the best image-based scores while FlexIT obtains the best similarity w.r.t the target instruction as indicated by CLIPScore. Knowing that the CLIPScore is heavily biased for models based on the CLIP model (as FlexIT does), and considering the image-based scores, DM-Align achieves the best trade-off between similarities to the input image and the target instruction.}
\label{tab:global_evaluation_07}
\begin{center}
{
\begin{tabular}{l  l  c c c c }
\hline
 \multicolumn{2}{}{} & {\textbf{FID$\downarrow$}} &  {\textbf{LPIPS$\downarrow$}} & {\textbf{PWMSE$\downarrow$}} & {\textbf{CLIPScore$\uparrow$}} \\
 \hline
Dream & FlexIT & 150.20  $\pm$ 0.67 & 0.53  $\pm$ 0.00 & 47.63  $\pm$ 0.13 & \textbf{0.87  $\pm$ 0.00} \\
 & InstructPix2Pix & 158.77  $\pm$ 3.03 & 0.44  $\pm$ 0.00 & 43.20  $\pm$ 0.44 & 0.81  $\pm$ 0.00 \\
 & ControlNet & 140.42  $\pm$ 0.38 & 0.49  $\pm$ 0.00 & 49.6  $\pm$ 0.46 & 0.80  $\pm$ 0.00 \\
 & DiffEdit & 126.77  $\pm$ 0.14 & \textbf{0.29  $\pm$ 0.57} & 30.22  $\pm$ 0.14 & 0.72  $\pm$ 0.00 \\
 & Plug-and-Play & 128.13  $\pm$ 0.98 & 0.53  $\pm$ 0.00 & 48.56  $\pm$ 0.13 & 0.76  $\pm$ 0.00 \\
 & Imagic & 157.06 $\pm$ 0.23 & 0.65 $\pm$ 0.00 & 48.23 $\pm$ 0.03 & 0.79 $\pm$ 0.00 \\
 & DM-Align & \textbf{124.57  $\pm$ 0.52} & 0.31  $\pm$ 0.00 & \textbf{29.24  $\pm$ 0.03} & 0.80  $\pm$ 0.00 \\
 \hline

 Bison & FlexIT & 41.78  $\pm$ 0.09 & 0.50  $\pm$ 0.00 & 42.59  $\pm$ 0.03 & \textbf{0.90  $\pm$ 0.00} \\
 & InstructPix2Pix & 62.62  $\pm$ 0.17 & 0.53  $\pm$ 0.00 & 41.45  $\pm$ 0.01 & 0.78  $\pm$ 0.00 \\
 & ControlNet & 45.87  $\pm$ 0.38 & 0.45  $\pm$ 0.00 & 52.12  $\pm$ 0.11 & 0.78  $\pm$ 0.00 \\
 & DiffEdit & 53.54  $\pm$ 0.22 & 0.45  $\pm$ 0.00 & 49.65  $\pm$ 0.18 & 0.76  $\pm$ 0.00 \\
  & Plug-and-Play & 52.44  $\pm$ 0.18 & 0.46  $\pm$ 0.00 & 48.45  $\pm$ 0.15 & 0.76  $\pm$ 0.00 \\
  & Imagic & 63.23 $\pm$ 0.28 & 0.52 $\pm$ 0.00 & 51.44 $\pm$ 0.12 &0.77 $\pm$0.00 \\
 & DM-Align & \textbf{40.05  $\pm$ 0.03} & \textbf{0.39  $\pm$ 0.00} & \textbf{37.05  $\pm$ 0.07} & 0.78  $\pm$ 0.00 \\
 
\hline
 Imagen & FlexIT & 91.86  $\pm$ 0.32 & 0.46  $\pm$ 0.00 & 44.05  $\pm$ 0.00 & \textbf{0.91  $\pm$ 0.00} \\
 & InstructPix2Pix & 133.33  $\pm$ 0.04 & 0.57  $\pm$ 0.00 & 42.68  $\pm$ 0.18 & 0.79  $\pm$ 0.00 \\
 & ControlNet & 85.86  $\pm$ 0.26 & 0.51  $\pm$ 0.00 & 58.44  $\pm$ 0.04 & 0.79  $\pm$ 0.00 \\
 & DiffEdit & 101.73  $\pm$ 0.00 & 0.38  $\pm$ 0.00 & 30.02  $\pm$ 0.09 & 0.71  $\pm$ 0.00 \\
 & Plug-and-Play & 84.37  $\pm$ 0.29 & 0.41  $\pm$ 0.00 & 41.79  $\pm$ 0.07 & 0.78  $\pm$ 0.00 \\
 & Imagic &  94.92  $\pm$ 0.44 & 0.67  $\pm$ 0.00 &   51.58  $\pm$ 0.11 & 0.77  $\pm$ 0.00 \\
 & DM-Align & \textbf{66.68  $\pm$ 0.01} & \textbf{0.31  $\pm$ 0.00} & \textbf{29.04  $\pm$ 0.01} & 0.79  $\pm$ 0.00 \\
\hline
\end{tabular} }
\end{center}
\end{table*}

\begin{table*}[t]
\small
\caption{
Image-level evaluation of DM-Align on a subset of the Bison dataset that contains only source and target text instructions with a degree of similarity higher than Rouge 0.7. Out of all baselines, only FlexIT and DiffEdit are presented, as they utilize a source caption in their implementation. While DM-Align scores better than the baselines for image-based metrics, FlexIT has the highest CLIPScore due to its CLIP-based architecture.}\label{tab:results_bison_07}
\begin{center}
{
\begin{tabular}{l c c c c  }
\hline
{} & {\textbf{FID$\downarrow$}} &  {\textbf{LPIPS$\downarrow$}} & {\textbf{PWMSE$\downarrow$}} & {\textbf{CLIPScore$\uparrow$}}\\
\hline
 FlexIT & 71.64  $\pm$ 0.03 & 0.48  $\pm$ 0.00 & 42.30  $\pm$ 0.03 & \textbf{0.89  $\pm$ 0.00}\\
 DiffEdit & 74.60  $\pm$ 0.94 & 0.44  $\pm$ 0.01 & 51.75  $\pm$ 0.29 & 0.76  $\pm$ 0.00\\
 DM-align & \textbf{67.91  $\pm$ 0.00} & \textbf{0.36  $\pm$ 0.00} & \textbf{36.28  $\pm$ 0.00} & 0.78  $\pm$ 0.00\\

 \hline
\end{tabular}   }
\end{center}

\end{table*}

\begin{table*}[t]
\small
 \caption{Background-level evaluation for Dream, Imagen and Bison datasets (mean and variance). DM-Align outperforms the baselines in terms of background preservation, especially for the Bison and Imagen datasets which have more elaborate captions than Dream.}
\label{tab:background_evaluation_07}
\begin{center}
{
\setlength{\tabcolsep}{4.5pt}
\begin{tabular}{l  l  c c c }
\hline
\multicolumn{2}{}{} & {\textbf{FID$\downarrow$}} &  {\textbf{LPIPS$\downarrow$}}  & {\textbf{PWMSE$\downarrow$}}\\
\hline

 Dream & FlexIT & 154.44  $\pm$ 0.19 & 0.31  $\pm$ 0.00 & 30.22  $\pm$ 0.05 \\
 & InstructPix2Pix & 147.62  $\pm$ 0.82 & 0.25  $\pm$ 0.00 & 27.87  $\pm$ 0.35 \\
 & ControlNet & 137.29  $\pm$ 1.86 & 0.31  $\pm$ 0.00 & 32.74  $\pm$ 0.42 \\
 & DiffEdit & 125.95  $\pm$ 0.44 & 0.15  $\pm$ 0.00 & 15.72  $\pm$ 0.04 \\ 
 & Plug-and-Play & 151.42  $\pm$ 1.02 & 0.34  $\pm$ 0.00 & 31.59  $\pm$ 0.00 \\
 & Imagic & 174.41  $\pm$ 1.49 & 0.42  $\pm$ 0.00 & 31.63  $\pm$ 0.08 \\
 & DM-Align & \textbf{102.44  $\pm$ 0.07} & \textbf{0.11  $\pm$ 0.00} & \textbf{14.54  $\pm$ 0.01} \\
 \hline
 Bison & FlexIT & 35.48  $\pm$ 0.07 & 0.24  $\pm$ 0.00 & 20.38  $\pm$ 0.03 \\
 & InstructPix2Pix & 44.01  $\pm$ 0.28 & 0.26  $\pm$ 0.00 & 20.00  $\pm$ 0.06 \\
 & ControlNet & 35.39  $\pm$ 0.06 & 0.25  $\pm$ 0.00 & 26.58  $\pm$ 0.04 \\
 & DiffEdit & 37.68  $\pm$ 0.33 & 0.23  $\pm$ 0.00 & 19.68  $\pm$ 0.09 \\
  & Plug-and-Play & 36.44  $\pm$ 0.78 & 0.24  $\pm$ 0.00 & 19.79  $\pm$ 0.12 \\
  & Imagic & 43.55  $\pm$ 0.76 & 0.27  $\pm$ 0.00 & 27.12  $\pm$ 0.10 \\
 & DM-Align & \textbf{16.41  $\pm$ 0.00} & \textbf{0.08  $\pm$ 0.00} & \textbf{14.16  $\pm$ 0.00} \\
\hline
 Imagen & FlexIT & 92.44  $\pm$ 0.35 & 0.36  $\pm$ 0.00 & 36.57  $\pm$ 0.01 \\
 & InstructPix2Pix & 124.32  $\pm$ 0.80 & 0.46  $\pm$ 0.00 & 34.29  $\pm$ 0.15 \\
 & ControlNet & 85.56  $\pm$ 0.31 & 0.42  $\pm$ 0.00 & 49.78  $\pm$ 0.02 \\
 & DiffEdit & 88.01  $\pm$ 0.55 & 0.31  $\pm$ 0.00 & 24.17  $\pm$ 0.09 \\
 & Plug-and-Play & 81.28  $\pm$ 0.28 & 0.34  $\pm$ 0.00 & 31.59  $\pm$ 0.07 \\
 & Imagic & 103.74  $\pm$ 1.49 & 0.56  $\pm$ 0.00 & 43.91  $\pm$ 0.08 \\
 & DM-Align & \textbf{54.12  $\pm$ 0.04} & \textbf{0.21  $\pm$ 0.00} & \textbf{22.09  $\pm$ 0.00} \\

% BISON & FlexIT & 56.62 $\pm$ 0.33 & 0.24 $\pm$ 0.24 & 20.24 $\pm$ 0.09  \\
% & DiffEdit & 63.44 $\pm$ 0.49 & 0.21 $\pm$ 0.03 & 25.76 $\pm$ 2.98  \\
% & ControlNet & 59.53 $\pm$ 0.85 & 0.18 $\pm$ 0.04 & 18.43 $\pm$ 1.96  \\
% & InstructPix2Pix & 63.93 $\pm$ 0.55 & 0.23  $\pm$ 0.18 & 20.35 $\pm$ 1.87 \\
% & Plug-and-Play & 41.45 $\pm$ 0.74 & 0.19 $\pm$ 0.49 & 21.53 $\pm$ 1.31 \\
% & DM-Align & \textbf{23.37 $\pm$ 0.98} & \textbf{0.05 $\pm$ 0.00} & \textbf{13.03 $\pm$ 0.39} \\
% \hline
% Imagen & FlexIT &  92.86  $\pm$   0.57  & 0.36  $\pm$ 0.48 & 36.35  $ \pm$ 0.48 \\
% & DiffEdit &  93.83  $\pm$   0.49  & 0.31  $\pm$ 0.44 & 23.1  $ \pm$ 0.44 \\
% & ControlNet & 75.34  $\pm$   0.85  & 0.33  $\pm$ 0.67 & 48.88  $ \pm$ 0.67 \\
% & InstructPix2Pix & 136.74  $\pm$   0.49  & 0.42  $\pm$ 0.45 & 36.11  $ \pm$ 0.45 \\
% & Plug-and-Play & 81.28  $\pm$   0.71  & 0.34  $\pm$ 0.76 & 34.73  $ \pm$ 0.76 \\
% & DM-Align & 54.14  $\pm$   0.7  & 0.21  $\pm$ 0.33 & 22.09  $ \pm$ 0.33 \\
% \hline
% Dream & FlexIT &  152.44  $\pm$   0.42  & 0.31  $\pm$ 0.91 & 30.25  $ \pm$ 0.91  \\
% & DiffEdit &  148.57  $\pm$   0.47  & 0.3  $\pm$ 0.63 & 33.45  $ \pm$ 0.63 \\
% & ControlNet & 139.91  $\pm$   0.69  & 0.29  $\pm$ 0.63 & 33.31  $ \pm$ 0.63 \\
% & InstructPix2Pix & 144.49  $\pm$   0.84  & 0.24  $\pm$ 0.54 & 28.32  $ \pm$ 0.54 \\
% & Plug-and-Play & 151.42  $\pm$   0.48  & 0.34  $\pm$ 0.47 & 31.59  $ \pm$ 0.47 \\
% & DM-Align & 52.35  $\pm$   0.82  & 0.09  $\pm$ 0.22 & 13.53  $ \pm$ 0.22    \\
\hline
\end{tabular} }
\end{center}

\end{table*}

\subsection{Diffusion mask} \label{sec:background}

To ensure the coherence of the complete image given the target language instruction and to cope with the cases when the object to be replaced is smaller than the object to be inserted, we also use a global diffusion mask. The computation of the diffusion mask represents the third step of our proposed model (Figure \ref{fig:overview}) and relies on the denoising diffusion probabilistic models (DDPM) \citep{DBLP:conf/nips/HoJA20, weng2021diffusion}. DDPMs are based on Markov chains that gradually convert the input data into Gaussian noise during a forward process, and slowly denoise the sampled data into newly desired data during a reverse process.
In each iteration $t$ of the forward process, new data $x_t$ is sampled from the distribution $q(x_t|x_{t-1}) = \mathcal{N}(\sqrt{1-\beta}x_{t-1}, \beta\mathit{I})$, where $\beta_t$ is an increasing coefficient that varies between $0$ and $1$ and controls the level of noise for each time step $t$. The process is further simplified by expressing the sampled data $x_t$ w.r.t the input image $x_0$, as follows:
\begin{equation}
x_t = \sqrt{\alpha_t}x_0+\sqrt{1-\alpha_t}\epsilon
\end{equation}
where $\alpha_t = \prod_{i=0}^t(1-\beta_i)$ and $\epsilon\sim\mathcal{N}(0,1)$ represents the noise variable. As we empirically observed that the editing effect is diminished over the regions where the noise variable is cancelled, we set the noise variable $\epsilon$ to 0 over the regions that should be preserved. We dubbed this operation noise cancellation. The forward process is executed for $T$ iterations until $x_T$ converges to $\mathcal{N}(0,1)$. 
%During the reverse process, new data is sampled from the distribution $q(x_{t-1}|x_t)$. As the distribution $q(x_{t-1}|x_t)$ is not easily computable, it is approximated by a model $p_\theta$ usually represented by a U-Net neural network. 
%During the reverse process, at each time step $t-1$, the data is denoised from the distribution $p_\theta(x_{t-1}|x_t) = \mathcal{N}(\sqrt{\alpha_{t-1}}x_0 + \sqrt{1-\alpha_{t-1}-\sigma^2_t}\frac{x_t - \sqrt{\alpha_t}x_0}{\sqrt{1-\alpha_t}})$, where $\sigma^2$ represents the variance. 
During the reverse process, at each time step $t-1$, $x_{t-1}$ is denoised from the distribution $q_\theta(x_{t-1}|x_t)$ defined as:
\begin{multline}\label{eq:reverse_process}
    q_\theta(x_{t-1}|x_t) = \mathcal{N}(\frac{1}{\sqrt{1-\beta_t}}(x_t- \\
    \frac{\beta_t}{\sqrt{1-\alpha_t}}\epsilon_{\theta}(x_t)), \frac{1-\alpha_{t-1}}{1-\alpha_t}\beta_t)
\end{multline}
where $\epsilon_{\theta}(x_t, t)$ is estimated by a neural network usually represented by a U-Net.

%\sqrt{\alpha_{t-1}}x_0 + \sqrt{1-\alpha_{t-1}-\sigma^2_t}\frac{x_t - \sqrt{\alpha_t}x_0}{\sqrt{1-\alpha_t}})$, where $\sigma^2$ represents the variance. 
% After the definition of the forward and reverse processes, the training of DDPM relies on the variational lower bound as follows:
% \begin{equation}
% \begin{array}{l}
% log(p(x_0) \geq log{p_\theta}(x_0|x_1)- D_{KL}(q(x_{1:T}|x_0)||p(x_{1:T}|x_0)) \\ = L_0 - \sum_{t=1}^T{L_{t}}
% \end{array}
% \end{equation}
% where $D_{KL}$ represents the Kullback–Leibler divergence, $L_0$ is the reconstruction loss, $L_T$ shows the proximity of $x_T$ to the Gaussian noise and $L_{t}$ (t = $\overline{1,T-1}$) indicates the closeness between the denoised step $p(x_{t}|x_{t+1})$ and the approximated one $q(x_{t}|x_{t+1})$.

To impose the text conditionally in a diffusion model, we have to integrate the text instruction $c$ into the U-Net model and compute $\epsilon_{\theta}(x_t|c)$, instead of $\epsilon_{\theta}(x_t)$. Using classifier-free guidance \citep{DBLP:journals/corr/abs-2205-11487} and knowing that $s$ ($s > 1$) represents the guidance scale, $\epsilon_{\theta}(x_t)$ mentioned in Eq. \ref{eq:reverse_process} is replaced by $\overline{\epsilon_{\theta}}(x_t|c)$ defined as:
\begin{equation}
\overline{\epsilon_{\theta}}(x_t|c) = s\epsilon_{\theta}(x_t|c) + (1-s)(\epsilon_{\theta}(x_t|0)
\end{equation}

To obtain the diffusion mask, we first compute the denoised output of the input image corresponding to the source instruction and the denoised output of the input image corresponding to the target instruction by running two separate DDPM processes. The diffusion process does not run over the input image but over its encoded representation yielded by a Variational Autoencoder (VAE) \citep{DBLP:journals/corr/KingmaW13, DBLP:conf/cvpr/RombachBLEO22} with Kullback-Leibler loss. Therefore, the denoised outputs do not represent the final edited image but only an intermediate image representation with semantic information associated with the source or target instruction. Inspired by \citet{DBLP:journals/corr/abs-2210-11427}, we compute the diffusion mask as the absolute difference between the two noise estimates that is rescaled between $[0,1]$ and binarized using a threshold set to $0.5$. This diffusion mask represents a global mask that roughly indicates the content to be changed.

%The noise estimates are generated using classifier-free guidance \citep{DBLP:journals/corr/abs-2205-11487}. Given the guidance scale $s$ ($s > 1$) and the outputs of the denoising step $t$ $\epsilon_{\theta}(x_t|t)$ and $\epsilon_{\theta}(x_t|t,c)$ unconstrained, respectively constrained by the text instruction $c$, the final noise estimate of the denoising step $t$ is:
% \begin{equation}
% \overline{\epsilon_{\theta}}(x_t|t, c) = s\epsilon_{\theta}(x_t|t,c) + (1-s)(\epsilon_{\theta}(x_t|t)
% \end{equation}

\subsection{Refinement of the diffusion mask} \label{sec: fusion}

The refinement of the diffusion mask represents the fourth step of DM-Align as presented in Figure \ref{fig:overview}. To further improve the precision of the global diffusion mask, we refine it using the regions detected in Section \ref{segmentation_image_word_align}. More specifically, we extend the diffusion mask to include the regions to be altered and shrink it to avoid editing over the preserved regions. To improve control over the preserved background, we adjust the noise variable over the forward process of the obtained diffusion mask. The noise variable is cancelled for the unaltered regions detected in the previous step and kept unchanged for the regions to be manipulated.

\begin{table*}[t]
\small
\caption{Ablation tests for the Imagen dataset (mean and variance). The results underscore the significance of all DM-Align components. "Non-shared objects" denote objects mentioned solely in the source caption, while "Refinement of diffusion mask" involves adjusting the diffusion mask through shrinkage or expansion based on regions corresponding to keywords.} 
\begin{center}
{
\begin{tabular}{l  c c c c  }
\hline
  \multicolumn{1}{}{} & {\textbf{FID$\downarrow$}} &  {\textbf{LPIPS$\downarrow$}} & {\textbf{PWMSE$\downarrow$}} & {\textbf{CLIPScore$\uparrow$}} \\
\hline
(w/o) diffusion mask & 43.36 $\pm$ 1.44 & 0.42 $\pm$ 0.00 & 41.61 $\pm$ 0.26 & 0.77 $\pm$ 0.00 \\
(w/o) noise cancellation & 44.44 $\pm$ 0.76 & 0.41 $\pm$ 0.00 & 40.57 $\pm$ 0.30 & \textbf{0.79 $\pm$ 0.00} \\
(w/o) refinement of diffusion mask & 47.63 $\pm$ 0.78 & 0.43 $\pm$ 0.00 & 43.60 $\pm$ 0.15 & 0.77 $\pm$ 0.00 \\
(w/o) objects with different modifiers & 42.34 $\pm$ 0.57 & 0.40 $\pm$ 0.00 & 38.23 $\pm$ 0.20 & 0.77 $\pm$ 0.00 \\
(w/o) non-shared objects & 45.35 $\pm$ 2.25 & 0.43  $\pm$ 0.00 & 41.57 $\pm$ 0.79 & 0.77 $\pm$ 0.00 \\
DM-Align & \textbf{40.05  $\pm$ 0.00} & \textbf{0.39  $\pm$ 0.00} & \textbf{37.05  $\pm$ 0.00} & 0.78  $\pm$ 0.00  \\
\hline
\end{tabular} }
\end{center}    
\label{tab:ablation_tests}
\end{table*}

Note that both the global diffusion mask with noise cancellation and the regions determined through image segmentation are necessary for a qualitative mask. The global diffusion mask facilitates the replacement of small objects with larger ones and gives context to the editing. On the other hand, the insertion or deletion of different regions based on image segmentation improves the precision of the final mask as shown in ablation experiments in Subsection \ref{qq_analysis}.  

Once the refined diffusion mask is computed, we use inpainting stable diffusion \citep{DBLP:conf/cvpr/RombachBLEO22} to edit the masked regions based on the given target text caption (fifth step of DM-Align presented in Figure \ref{fig:overview}). We also tried to replace the inpainting stable diffusion with latent blended diffusion \citep{DBLP:journals/corr/abs-2206-02779}. However, the obtained results were slightly worse, and the computational time increased by 60\% (details are in Table 6 of Appendix C).

\section{Experimental setup}\label{experiments}

\textbf{Baselines.} We compare results obtained with DM-Align with those of FlexIT \citep{couairon2022flexit}, DiffEdit \citep{DBLP:journals/corr/abs-2210-11427}, ControlNet \citep{DBLP:journals/corr/abs-2302-05543}, Plug-and-Play \citep{DBLP:conf/cvpr/BrooksHE23}, Imagic \citep{DBLP:conf/cvpr/KawarZLTCDMI23} and IntructPix2Pix \citep{DBLP:conf/cvpr/TumanyanGBD23}. The implementation details are presented in Appendix A.

\begin{figure*}[t]
\centering
   {   \includegraphics[width=14cm]
   {"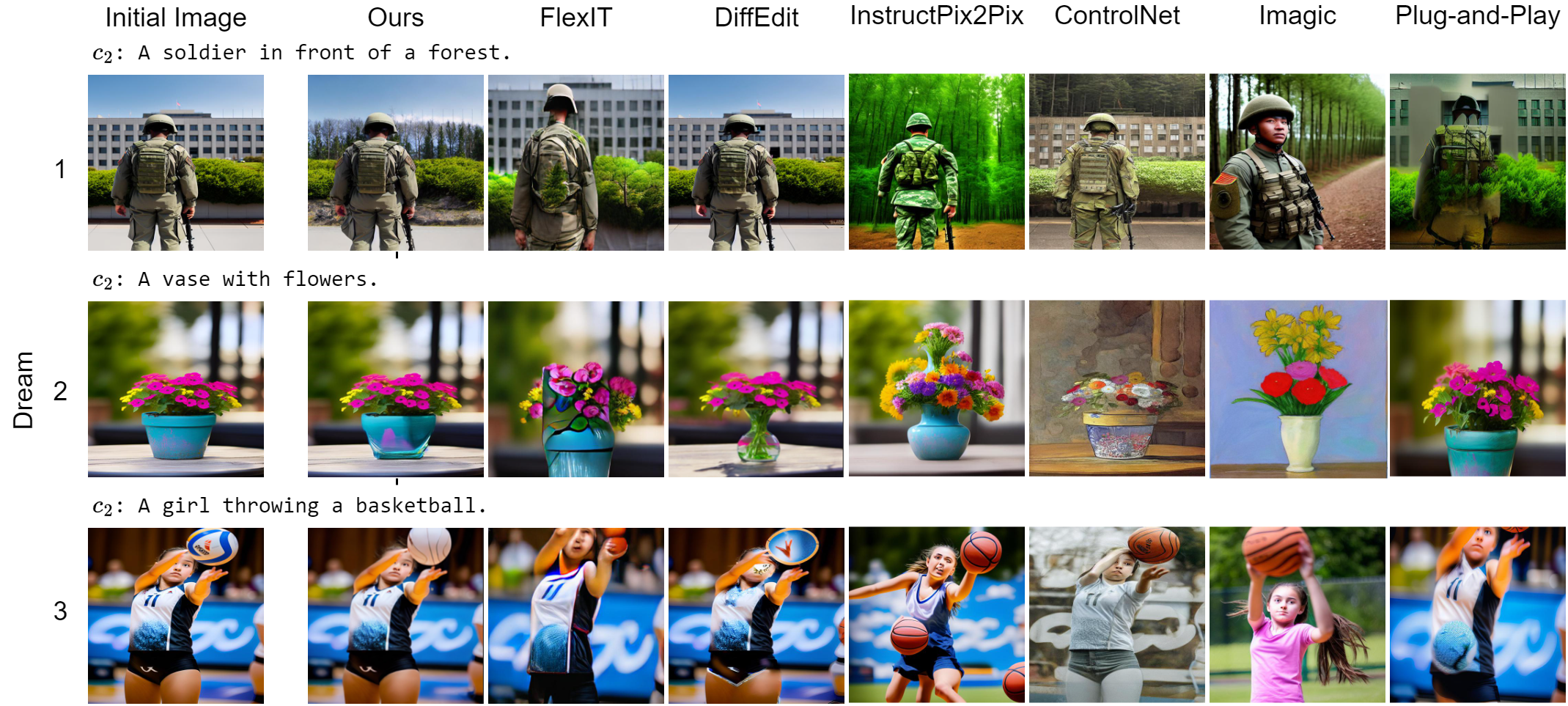"}}
  { \caption{Semantic image editing: Dream dataset. Source captions: (1) $c_1$. A soldier in front of a building. (2) $c_1$. A pot with flowers. (3) $c_1$. A girl throwing a volleyball.
   }}
   
\label{fig:examples_dream}
\end{figure*}

\textbf{Datasets.} While ControlNet, Plug-and-Play, Imagic and InstructPix2Pix are evaluated on datasets devoid of source text descriptions, FlexIT and DiffEdit are evaluated on a subset of the ImageNet dataset \citep{DBLP:conf/cvpr/DengDSLL009}, which assumes the replacement of the main object of the image with another object. Additionally, DiffEdit is evaluated on the Bison dataset \citep{DBLP:conf/iccvw/HuMM19} and a self-defined collection of Imagen pictures \citep{DBLP:journals/corr/abs-2205-11487}. Out of these datasets, our model is evaluated using the Bison dataset and the collection of images generated by Imagen (further referred to as the Imagen dataset) described by \citet{DBLP:journals/corr/abs-2210-11427}. We omit the ImageNet dataset due to its oversimplified setup, primarily employing single-word source and target text instructions.

Bison and Imagen datasets contain elaborated text captions with up to 23 words. To investigate the behavior of the DM-Align model and the baseline models when confronted with shorter text instructions we generate a collection of $100$ images using Dream by WOMBO\footnote{The code is available at \url{https://github.com/cdgco/dream-api}} that relies on the source captions as guidance. As the dataset is generated using Dream by Wombo, we further refer to it as Dream. To complete the Dream dataset, we specify a new text query as the target instruction for each image-instruction pair. Unlike the Imagen and Bison datasets, the text instructions of Dream do not contain more than 11 words.

\textbf{Evaluation metrics.}

To evaluate our model, we use a set of metrics that assess the similarity of the edited image to both the input image and the target instruction. By default, it is a trade-off between image-based and text-based metrics as we need to find the best equilibrium point. 

Generating images close to the source image improves the image-based metrics while reducing the similarity to the target caption. On the other hand, images close to the target instruction improve the text-based scores but can affect the similarity to the input picture. The equilibrium point is important given that people tend to focus mainly on specifying the desired changes in an image while omitting the information that already exists \citep{hurley2014concise}. Therefore, the edited content can represent a small region of the new image while the rest of it should keep the content of the source image.

The similarity (or the distance) of the updated image w.r.t the source image is assessed using FID \citep{DBLP:conf/nips/HeuselRUNH17}, LPIPS \citep{DBLP:conf/cvpr/ZhangIESW18} and the pixel-wise Mean Square Error (PWMSE). FID relies on the difference between the distributions of the last layer of the Inception V3 model \citep{DBLP:conf/cvpr/SzegedyVISW16} that separately runs over the input and edited images. FID measures the consistency and image realism of the new image w.r.t the source image. Contrary to the quality assessment computed by FID, LPIPS measures the perceptual similarity by calculating the distance between layers of an arbitrary neural network that separately runs over the input and updated images. As the LPIPS metric, PWMSE determines the pixel leakage by computing the pixel-wise error between the input and the edited images. The similarity of the updated image w.r.t the target instruction is computed in the CLIP multimodal embedding space by the CLIPScore \citep{DBLP:conf/emnlp/HesselHFBC21}. More details about the evaluation metrics are specified in Appendix B.

% \begin{table*}[t]
% \caption{Evaluation of the DM-Align rules: D$_1$ is the subset of the BISON$_{0.6}$ dataset that requires , D$_2$ is the subset of the BISON$_{0.7}$ dataset that requires the 2nd rule (mean and variance). All metrics indicate the relevance of the two additional rules.} \label{tab:ablation_tests_subset}
% \begin{center}
% {\footnotesize
% \begin{tabular}{l | l | c c c c  }
% \multicolumn{2}{}{} & {FID$\downarrow$} &  {LPIPS$\downarrow$} &  {PWMSE$\downarrow$} & {CLIPScore$\uparrow$}  \\
% D$_1$ & (w/o) 1st rule & 104.451 $\pm$ 4.001 & 0.339 $\pm$ 0.002 & 35.256 $\pm$ 0.763 & 0.762 $\pm$ 0.001  \\
% & (w/) 1st rule & \textbf{96.002 $\pm$ 1.997} & \textbf{0.332 $\pm$ 0.003} & \textbf{32.379 $\pm$ 0.220} & \textbf{0.782 $\pm$ 0.004}\\
% D$_2$ & (w/o) 2nd rule & 94.414 $\pm$ 0.122 & 0.271 $\pm$ 0.001 & 34.211 $\pm$ 0.864 & 0.762 $\pm$ 0.002  \\
% & (w/) 2nd rule & \textbf{85.003 $\pm$ 1.24} & \textbf{0.259 $\pm$ 0.000} & \textbf{34.198 $\pm$ 0.431} & \textbf{0.760 $\pm$ 0.002} \\
% \end{tabular} 
% }
% \end{center}
% \end{table*}

\section{Results and discussion}
\subsection{Quantitative analysis and ablation tests}\label{qq_analysis}

\noindent{\textit{How well can the DM-Align model
edit a source image considering the length of the text instruction?} To address the first research question, we refer to Table \ref{tab:global_evaluation_07}. When compared to the baselines Diffedit, ControlNet, FlexIT, Plug-and-Play, and InstructPix2Pix, our proposed DM-Align model exhibits particularly effective performance in terms of image-based metrics.}

This effectiveness is particularly noticeable in the Bison and Imagen datasets, which contain longer captions compared to the Dream dataset. When compared with the best baseline over the Imagen dataset, DM-Align improves FID, LPIPS, and PWMSE by 23.42\%, 19.87\%, and 3.32\%, respectively. Similar results are observed for the Bison dataset, where DM-Align enhances the results of the best baseline by 4.22\% for FID, by 14.26\% for LPIPS, and by 11.20\% for PWMSE. In the case of the Dream dataset, DM-Align still outperforms other baselines in terms of FID and PWMSE, albeit with smaller margins. However, in terms of LPIPS, DiffEdit outperforms DM-Align over the instance of the Dream dataset.

Given the results presented in Table \ref{tab:global_evaluation_07}, we posit that baselines find it easier to accurately edit images using short text instructions. Conversely, when text instructions are more elaborate, such as in the Bison and Imagen datasets, results significantly surpass those achieved by the baselines. DM-Align leverages word alignments between source and target instructions, highlighting their crucial role in facilitating effective image editing.

In terms of text-based metrics, CLIPScore suggests that FlexIT generates images closest to the target instructions. This outcome is likely attributed to FlexIT's architecture, which is based on a CLIP model—the same model used to calculate CLIPScore. This issue is highlighted in \citep{poole2022dreamfusion}. Another possible explanation is that FlexIT is trained to maximize the similarity between input images and instructions. As depicted in Figures 3, 5, and 6, FlexIT may sacrifice image quality for higher similarity scores. Regarding CLIPScore, DM-Align consistently outperforms Plug-and-Play, Imagic and DiffEdit baselines or is equally effective as InstructPix2Pix and ControlNet. DM-Align also outperforms Prompt-to-Prompt \citep{DBLP:journals/corr/abs-2208-01626}. As Prompt-to-Prompt can edit only self-generated images the comparison with DM-Align is limited only to text-based metrics like CLIPScore. More details about this comparison are presented in Appendix C.

Given the text-based and image-based metrics, DM-Align seems to properly preserve the content of the input image and obtain a better trade-off between closeness to the input picture and target instruction than the baselines.

While the above analysis demonstrates that elaborate text instructions do not affect the editing capabilities of DM-Align, unlike the baselines, we are also interested in examining how the degree of overlap between source and target captions impacts the quality of the edited image. To analyze this, we select 575 Bison instances with a similarity between the source and target instructions higher than Rouge 0.75. We do not conduct this analysis for the Imagen and Dream datasets as their text instructions already exhibit a level of similarity higher than Rouge 0.75. Our analysis is limited to DM-Align, FlexIT, and DiffEdit, as the other baselines do not utilize source captions in their implementation and are therefore omitted from this analysis. The results are presented in Table \ref{tab:results_bison_07}. We observe that while the results are similar to the image-based and text-based scores reported in Table \ref{tab:global_evaluation_07} for Bison, all models report better performance and an improved trade-off between image and text-based metrics. These results suggest that increased overlap between source and target captions enhances the quality of image editing.

\noindent{\textit{How well does the DM-Align model preserve the background?} To extract the background, we consider the DM-Align mask obtained after adjusting the diffusion mask. Upon analyzing the results presented in Table \ref{tab:background_evaluation_07}, the first notable observation is the significant reduction in the FID score of the DM-Align model by 73.27\% for the Bison dataset, 40.12\% for the Imagen dataset, and 20.58\% for the Dream dataset when compared with the best baseline. Similarly, the LPIPS and PWMSE scores also indicate significant margin reductions, particularly for the Bison and Imagen datasets. Concerning the Dream dataset, DM-Align still outperforms the best-performing baseline with a margin of 25.02\% for LPIPS and 7.80\% for PWMSE. While DM-Align consistently demonstrates superior results for background preservation, we infer that the baselines are relatively adept at preserving the background only when the instructions are short and simple, as observed in the case of the Dream dataset. This conclusion is further supported by the results presented in Table \ref{tab:global_evaluation_07}.}

% \begin{figure*}[t]

%   { \caption{Semantic image editing using BISON$_{0.7}$ and Dream datasets. \textbf{BISON$_{0.7}$ dataset}: (1) $c_2$. A man standing next to his elephant on the beach. (2) $c_2$. A vase filled with lots of colorful flowers. (3) $c_2$. A man eating a hot dog at a crowded event. (4) $c_2$. A plate of fruit next to a glass of milk. \textbf{Dream dataset}: (5) $c_2$. A girl throwing a basketball. (6) $c_2$. A vase with flowers. (7) $c_2$. A quattro formaggi pizza on a plate. (8) $c_1$. $c_2$. An owl sitting on an iron gate.
%    }}
%    {   \includegraphics[width=\linewidth]
%    {"Figures/image_collection_part.png"}}
% \label{fig:examples_bison_data}
% \end{figure*}

\paragraph{Ablation tests}
To run the ablation tests for DM-Align we rely on the Imagen dataset. According to Table \ref{tab:ablation_tests}, the absence of the refinement of the diffusion mask using the regions detected with the word alignment model and the Grounding-SAM segmentation model has the highest negative impact over the similarity w.r.t the input picture. As expected, a significant negative effect over the similarity with the input image is also noticed when omitting the deleted nouns or the nouns with different modifiers in the two queries. Similarly, noise cancellation and especially the diffusion mask also affect the conservation of the background. 
Including all the components in the architecture of DM-Align mainly facilitates the preservation of the input image and does not result in a reduction of the CLIPScore. Therefore, the inclusion of all these components in the DM-Align represents the best trade-off w.r.t the similarity to the input image and to the target caption.

\begin{figure}[t]
\begin{center}
   \includegraphics[width=\linewidth]{"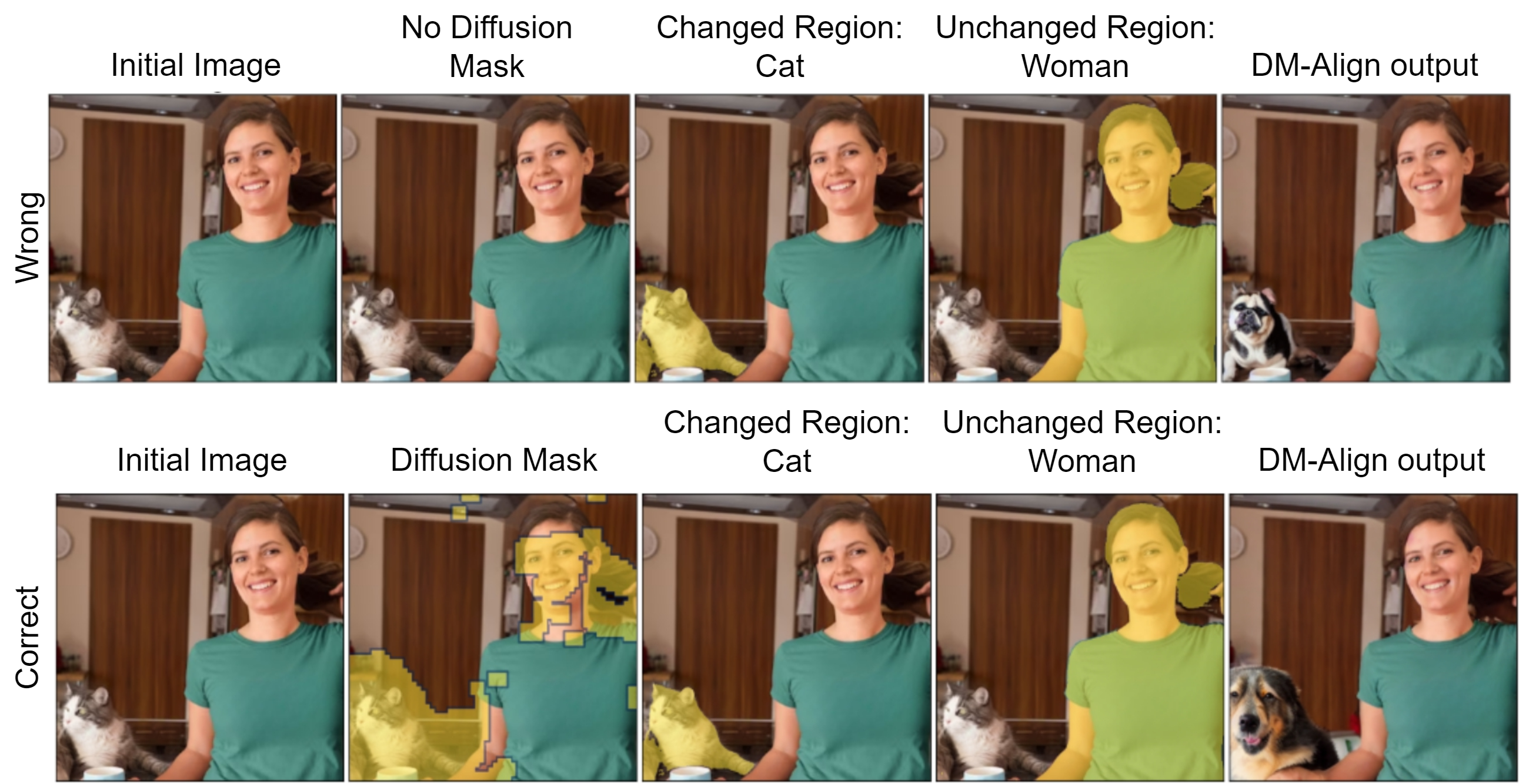"}
\end{center}
% \vspace{-0.5cm}
   \caption{1st line: Example of omitting the diffusion mask ($c_1$: A woman near a cat., $c_2$: A woman near a dog.). 2nd line: The correct example of including the diffusion mask.}
\label{fig:diffusion}
\end{figure}

The next five visualizations exemplify the ablation tests. The first row of each figure presents the effect of omitting a component of DM-Align, while the correct behavior is shown in the second row. Figure \ref{fig:diffusion} illustrates the effect of defining the editing mask based only on the image regions of the keywords. Without the diffusion mask, the model has to insert a new object in the fixed area of the replaced object. If we need to replace an object with a larger one, DM-Align without diffusion might create distorted and unnatural outputs. As we usually expect bigger dogs than cats, DM-Align with diffusion properly replaces the cat with a slightly bigger dog. On the contrary, the dog that replaced the cat is distorted when diffusion is not used. 

\begin{figure}[t]
\begin{center}
   \includegraphics[width=\linewidth]{"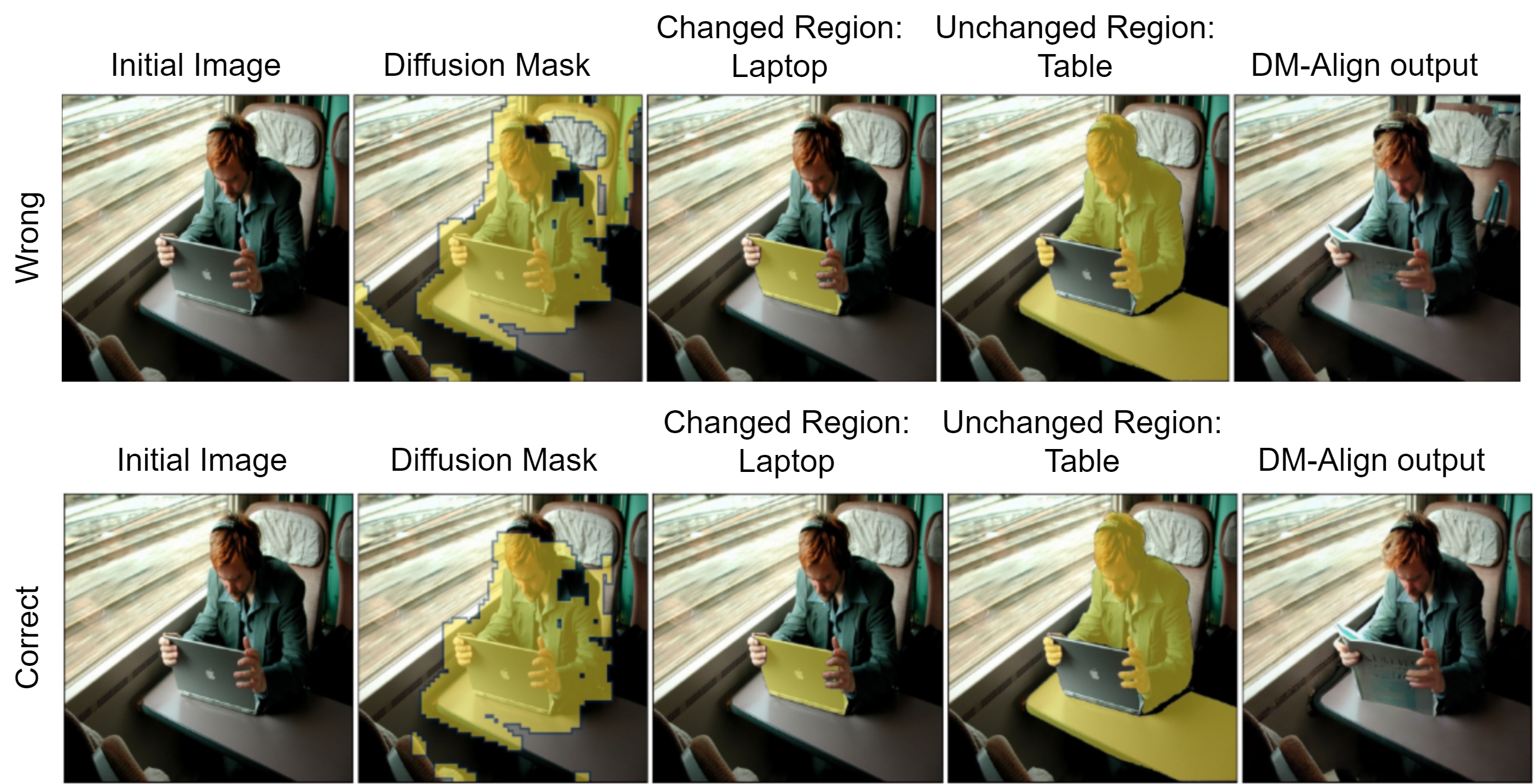"}
\end{center}
% \vspace{-0.5cm}
   \caption{1st line: Example of omitting the cancellation of the noise variable defined within the diffusion model. ($c_1$: A man sitting at a table holding a laptop on the train., $c_2$: A man sitting at a table reading a book on the train.). 2nd line: The correct example of including the noise cancellation.}
\label{fig:noise}
\end{figure}

\begin{figure}[t]
\begin{center}
   \includegraphics[width=\linewidth]{"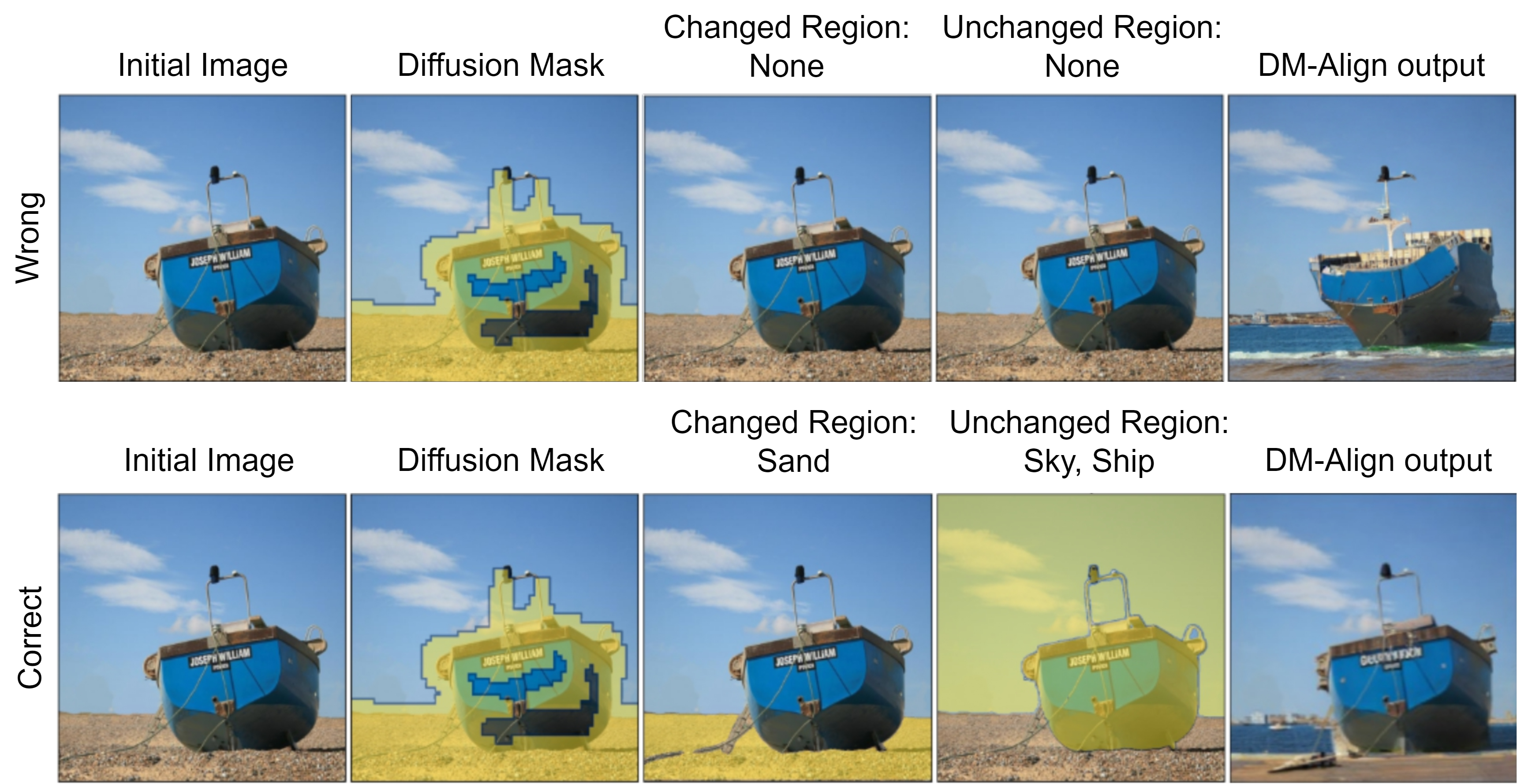"}
\end{center}
% \vspace{-0.5cm}
   \caption{1st line: Example of omitting the refinement of the diffusion mask using image segmentation ($c_1$: A clear sky and a ship landed on the sand., $c_2$: A clear sky and a ship landed on the ocean.). 2nd line: The correct example of including the refinement of the diffusion mask with image segmentation.}
\label{fig:segmentation}
\end{figure}

\begin{figure}[t]
\begin{center}
   \includegraphics[width=\linewidth]{"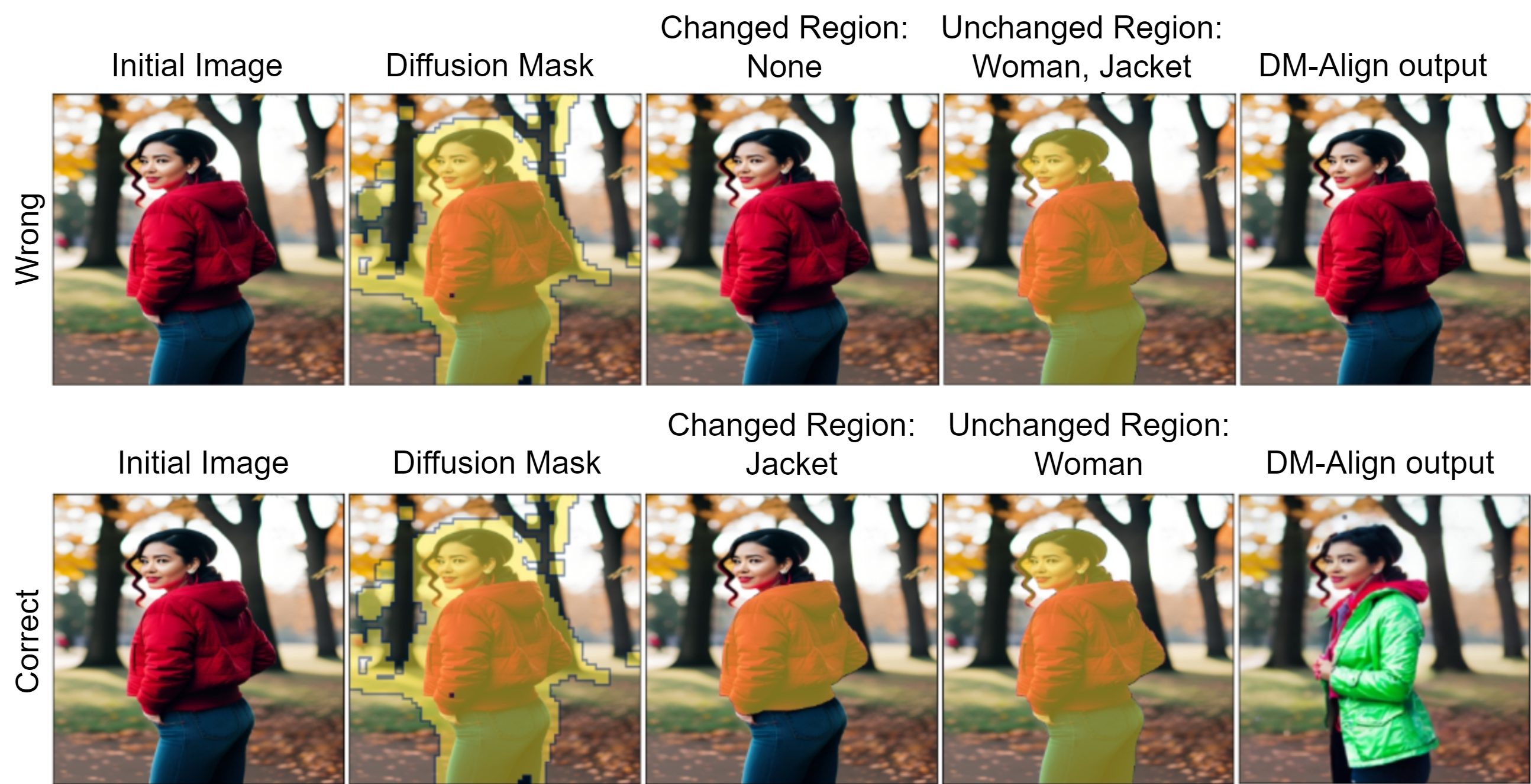"}
\end{center}
% \vspace{-0.5cm}
   \caption{1st line: Example of omitting the information about modifiers associated with the nouns shared by both captions ($c_1$: A woman with a red jacket., $c_2$:  A woman with a green jacket.). 2nd line: The correct example of including the information about the modifiers.}
\label{fig:modifier}
\end{figure}

\begin{figure}[t]
\begin{center}
   \includegraphics[width=\linewidth]{"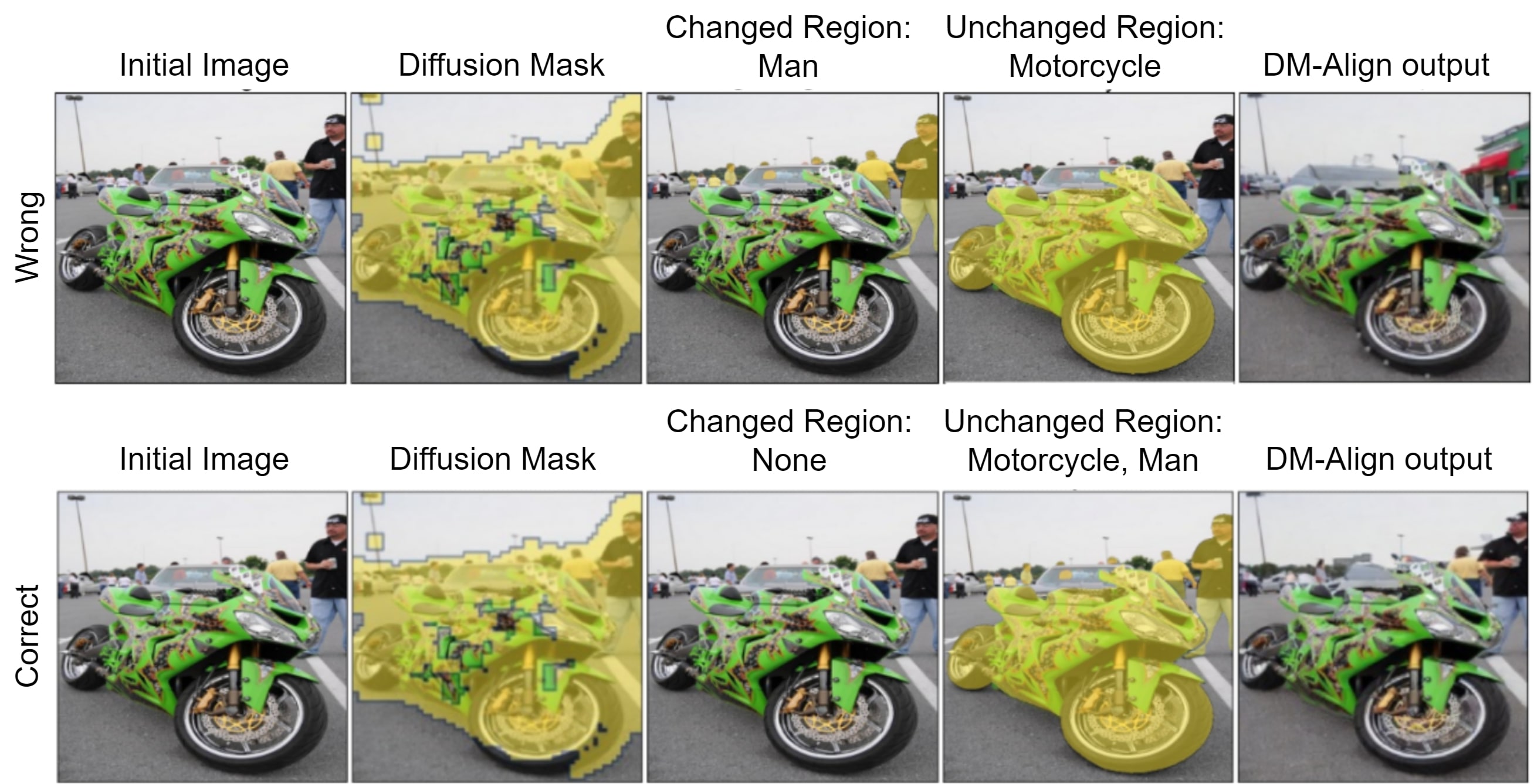"}
\end{center}
% \vspace{-0.5cm}
   \caption{1st line: Example of omitting the information about the deleted nouns from the source caption ($c_1$: A motorcycle near a man., $c_2$:  A motorcycle.). 2nd line: The correct example of including the information about the deleted nouns.}
\label{fig:delete}
\end{figure}

While the overall diffusion mask can give more context for the editing and allows the insertion of objects of different sizes, noise cancellation is an important step used to improve the initial diffusion mask. As shown in Figure \ref{fig:noise}, when noise cancellation is used, the initial diffusion mask is better trimmed, and the background is properly preserved.

As the diffusion mask does not have complete control over the regions to be edited, its extension or shrinkage based on the image regions of the keywords is mandatory to obtain a correct mask for editing. When the image is edited using only the initial diffusion mask in Figure \ref{fig:segmentation}, both the ship and the sand are modified, while the former is expected to be preserved. As opposed, when the diffusion mask is refined with image segmentation, only the sand is replaced by the ocean.

The omission of the adjective modifiers in the analysis of DM-Align is exemplified in Figure \ref{fig:modifier}. If the modifiers are left out, DM-Align considers the jacket a shared noun, like the noun ``woman", and removes its regions from the diffusion mask. As a result, DM-Align does not detect any semantical difference between the text instructions, and the output image is identical to the input image. On the other hand, if the modifiers are considered, DM-Align can properly adjust the color of the jacket while keeping the woman's face unaltered.

As we are interested to make only the necessary updates in the picture, while keeping the background and the regions of the deleted words unchanged, the region assigned to the word ``man" in Figure \ref{fig:delete} is removed from the diffusion mask. As a result, the corresponding region is untouched. On the contrary, the inclusion of the region associated with the word ``man" in the diffusion mask increases the randomness in the new image by inserting a store. Since the store is irrelevant, both the similarity scores w.r.t the input image or target instruction are reduced.

\subsection{Human qualitative analysis}

Some qualitative examples extracted from the three data collections are shown in Figures 3, 5, and 6. Compared to DIFFEdit, ControlNet, and FlexIT, as well as Plug-and-Play, Imagic and InstructPix2Pix, the DM-Align model demonstrates superior manipulation of the content of the input image while largely preserving the background in line with the target query. DM-Align establishes semantic connections between source and target queries, updating the image content accordingly, whereas the baselines often alter the background more than necessary, as discussed above. DiffEdit tends to introduce random visual content (see Figure 3), while FlexIT tends to distort and zoom into the image (Figures 5 and 6), trading off the minimization of the reconstruction loss with respect to the input image and the text instructions for potential distortions in the new image. Although ControlNet can maintain the structure of the input image, it struggles to preserve the texture or colors of the objects, likely due to the absence of a masking system. InstructPix2Pix also encounters challenges in preserving the style of objects in the input image and tends to include more objects in the image than specified in the target text instruction. Plug-and-Play zooms into the image and tends to slightly alter the details of objects requested for preservation in the target text instruction. Out of all baselines, Imagic shows the highest tendency to change the input image's compositional structure, as highlighted also by the image-based metrics presented in Tables \ref{tab:global_evaluation_07} and \ref{tab:background_evaluation_07}.

\begin{table}[h]
\small
\caption{Human evaluation of the quality of the editing process based on the text instruction (Q1), the preservation of the background (Q2) and the quality of the edited image (Q3). The results represent the average scores reported by annotators using a 5-point Likert scale.}
\label{tab:humna_evaluation}
\begin{center}
{
\begin{tabular}{l c c c  }
\hline
 & {\textbf{Q1$\uparrow$}} & {\textbf{Q2$\uparrow$}} & {\textbf{Q3$\uparrow$}} \\
 \hline
FlexIt & 3.75 & 4.00 & 3.85 \\
DiffEdit & 3.85 & 4.15 & 3.85\\
ControlNet & 3.50 & 3.75 & 3.90 \\
% Prompt-to-Prompt & 2.24 & 1.98 & 2.18 \\
Plug-and-Play & 3.80 & 4.10 & 3.85 \\
InstructPix2Pix & 3.50 & 3.75 & 3.80 \\
Imagic & 3.80 & 3.20 & 3.85 \\
DM-Align & \textbf{3.90} & \textbf{4.35} & \textbf{3.95}\\
\hline
\end{tabular} }
\end{center}
\vspace{-0.3cm}
\end{table}

To confirm the above observations, we randomly selected 100 images from the Bison dataset and asked Amazon MTurk annotators to evaluate the editing quality of the five baselines and the proposed DM-Align. For each edited image, the annotators were asked to evaluate the overall quality of the editing process based on the text instruction (Q1), the preservation of the background (Q2) and the quality of the edited image in terms of compositionality, sharpness, distortion, color and contrast (Q3). According to the human evaluation executed on a 5-point Likert scale, our model scores better than all baselines (Table \ref{tab:humna_evaluation}). The inter-rater agreement is good with Cohen's weighted kappa $\kappa$ between 0.65 and 0.75 for all analyzed models. 

\section{Conclusion, limitations and future work}\label{conclusion}

We propose a novel model DM-Align for semantic image editing that confers to the users a natural control over the image editing by updating the text instructions. By automatically identifying the regions to be kept or altered purely based on the text instructions, the proposed model is not a black box. Due to the high level of explainability, the users can easily understand the edited result and how to change the instructions to obtain the desired output.

The quantitative and qualitative evaluations show the superiority of DM-Align to enhance the text-based control of semantic image editing over existing baselines FlexIT, DiffEdit, ControlNet, Imagic, Plug-and-Play and InstructPix2Pix. Unlike the latter models, our approach is not limited by the length of the text instructions. Due to the inclusion of one-to-one alignments between the words of the instructions that describe the image before and after the image update, we can edit images regardless of how complicated and elaborate the text instructions are. Besides the low sensitivity to the complexity of the instructions, the one-to-one word alignments allow us to properly conserve the background while editing only what is strictly required by the users. 

DM-Align focuses on the editing of objects mentioned as nouns and their adjectives. In future work, its flexibility can be improved by editing actions in which objects and persons are involved. As a result, they might change position in the image without the need to update their properties.

\section*{Acknowledgments}
This project was funded by the European Research Council
(ERC) Advanced Grant CALCULUS (grant agreement No. 788506).

\bibliographystyle{acl_natbib}
\bibliography{custom}

\begin{figure*}[h]
\centering
   {   \includegraphics[width=12cm]
   {"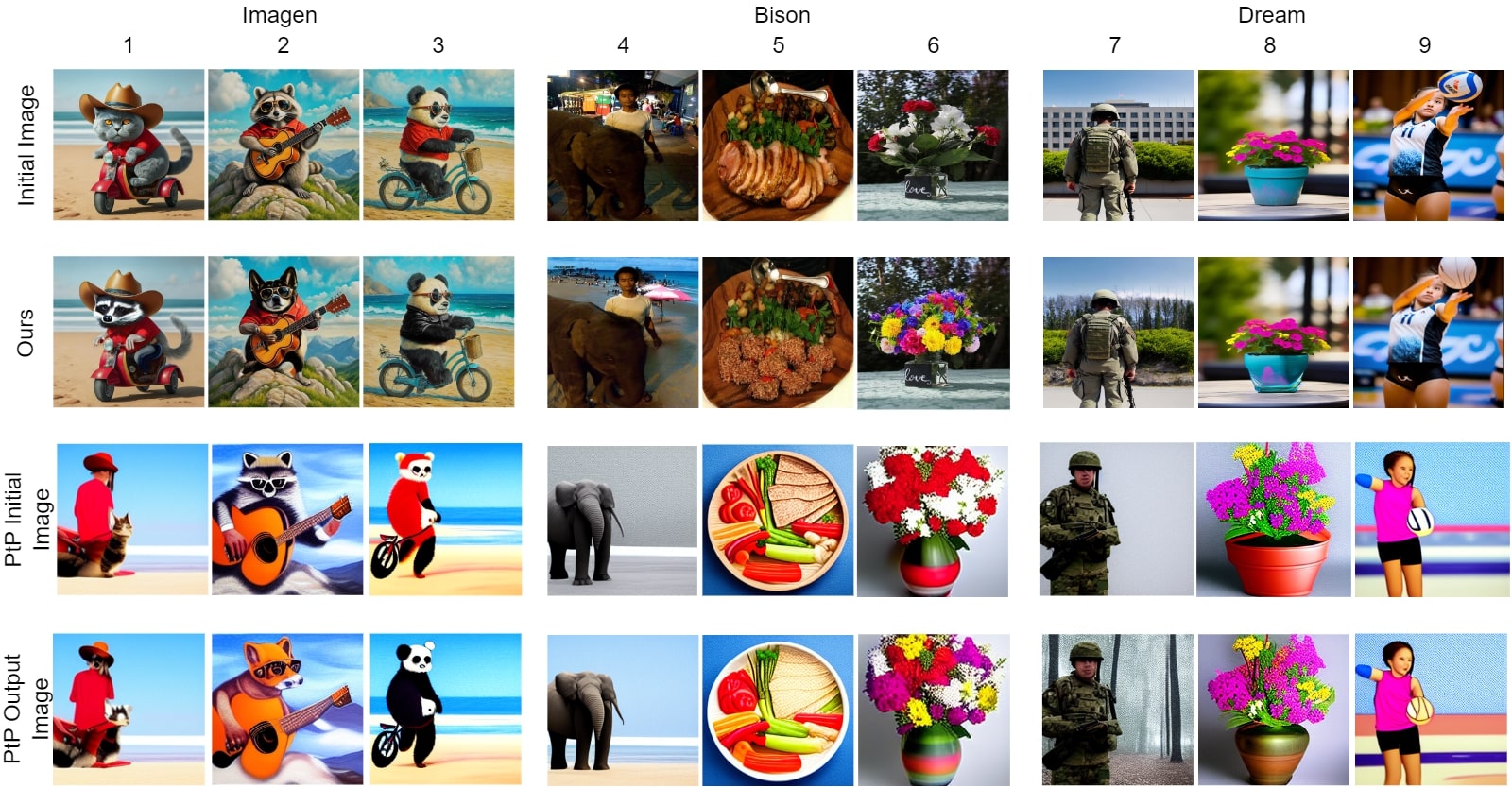"}}
   \caption{Semantic image editing: Comparison between Prompt-to-Prompt (PtP) and DM-Align. Source captions: (1) $c_1$. A photo of a British shorthair cat wearing a cowboy hat and red shirt riding a bike on a beach. $c_2$. A photo of a raccoon wearing a cowboy hat and red shirt riding a bike on a beach. (2) $c_1$. An oil painting of a raccoon wearing sunglasses and red shirt playing a guitar on top of a mountain. $c_2$. An oil painting of a Shiba Inu dog wearing sunglasses and red shirt playing a guitar on top of a mountain. (3) $c_1$. An oil painting of a fuzzy panda wearing sunglasses and red shirt riding a bike on a beach. $c_2$. An oil painting of a fuzzy panda wearing sunglasses and black jacket riding a bike on a beach. (4) $c_1$. A man standing next to a baby elephant in the city. $c_2$. A man standing next to his elephant on the beach. (5) $c_1$. A wooden plate topped with sliced meat and vegetables. $c_2$. A plate of rice meat and vegetables. (6) $c_1$. A vase filled with red and white flowers. $c_2$. A vase filled with lots of colorful flowers. (7) $c_1$. A soldier in front of a building. $c_2$. A soldier in front of a forest. (8) $c_1$. A pot with flowers. $c_2$. A vase with flowers. (9) $c_1$. A girl throwing a volleyball. $c_2$. A girl throwing a basketball.
   }\label{fig:examples_ptp}

\end{figure*}

\begin{table*}[h]
\small
\caption{Image-level evaluation of DM-Align with Stable diffusion and Blended latent diffusion for inpainting. The results are reported for the Dream dataset (mean and variance). }\label{tab:results_inpainting}

\begin{center}
{
\begin{tabular}{l c c c c  }
\hline
{} & {\textbf{FID$\downarrow$}} &  {\textbf{LPIPS$\downarrow$}} & {\textbf{PWMSE$\downarrow$}} & {\textbf{CLIPScore$\uparrow$}}\\
\hline
 DM-Align (Blended Latent Diffusion) & 128.87 $\pm$ 0.12 & 0.33 $\pm$ 0.00 & 32.50 $\pm$ 0.43 & 0.78 $\pm$ 0.00  \\
 DM-Align (Statble Latent Diffusion) & \textbf{124.57  $\pm$ 0.52} & \textbf{0.31  $\pm$ 0.00} & \textbf{29.24  $\pm$ 0.03} & \textbf{0.80  $\pm$ 0.00} \\
 \hline
\end{tabular}   }
\end{center}
\end{table*}

\begin{table}[ht]
\small
\caption{Image-level evaluation (mean and variance) of DM-Align and Prompt-to-Prompt using Dream and Imagen datasets.}\label{tab:ptp_clip}

\begin{center}
{
\begin{tabular}{l c c }
\hline
  &  {\textbf{CLIPScore$\uparrow$}} & {\textbf{CLIPScore$\uparrow$}} \\
  &  {(Dream)} & {(Imagen) } \\
 \hline
Prompt-to-Prompt & 0.76 $\pm$ 0.00 & 0.75 $\pm$ 0.00 \\
 DM-Align &  \textbf{0.80  $\pm$ 0.00} & \textbf{0.78  $\pm$ 0.00}\\
 \hline

\end{tabular} }
\end{center}
\end{table}

\section*{Appendix}

\section*{A. Implementation Details}\label{sec:implementation_details}
We use Stable Diffusion v1.4 with 50 diffusion steps and a guidance scale of 7.5 to run DM-Align. The size of the generated images is 512 $\times$ 512 pixels. For consistency reasons, we use Stable Diffusion v1.4 to run DiffEdit, ControlNet, Plug-and-Play and Imagic. In the case of InstructPix2Pix, we use the Stable Diffusion model fine-tuned by the authors. Following the implementation details mentioned in their papers, ControlNet, Plug-and-Play, DiffEdit, Imagic and InstructPix2Pix edit images at a resolution of 512 $\times$ 512 pixels. DiffEdit, ControlNet, Imagic and Plug-and-Play require 50 diffusion steps, while InstructPix2Pix manipulates images using 100 diffusion steps. The resolution of images edited by FlexIT is 288 $\times$ 288 pixels, according to the indication in its paper. All experiments are implemented using one NVIDIA GeForce RTX 3080 GPU.

\section*{B. Evaluation Metrics}\label{sec:eval_metrics}

Image-based evaluation metrics:
\begin{itemize}
    \item The FID score relies on the distribution of the output generated by the last layer of the Inception V3 model (Szegedy et al. 2016). %\cite{DBLP:conf/cvpr/SzegedyVISW16}%. 
    The metric is computed by measuring the Frechet distance between the distributions gleaned by running the Inception V3 model over the source and target images. Considering the mean $\mu_1$ and the covariance $C_1$ of the source images and the mean $\mu_2$ and the covariance $C_2$ of the target images, the FID score is computed as follows:
    \begin{equation}
    \begin{array}{l}
    FID = {|\mu_1 - \mu_2|}_2^2 + \\Tr(C_1+C_2-2{(C_1C_2)}^{1/2})
    \end{array}
    \end{equation}
    \item  LPIPS measures the average Euclidean distance between outputs of different layers of a neural network (AlexNet for the current study, as suggested by Zhang et al. (2018)) %\cite{DBLP:conf/cvpr/ZhangIESW18})%
    obtained by giving as input the source and the target images. Considering $x_1^l, \hat{x}_2^l \in \mathcal{R}^{H_l\times W_l \times C_l}$ as the intermediate $l$-th representations of the AlexNet for the source and the predicted target image, respectively, the LPIPS score is defined by:
    \begin{equation}
    \begin{array}{l}
    LPIPS = \sum_l{\frac{1}{H_l W_l} \sum_{h,w} {{|{x_1}_{hw}^l - (\hat{x}_2)_{hw}^l|}_2^2}}
    \end{array}
    \end{equation}
    \item PWMSE measures the pixel-wise mean square error between the input and the edited image.
\end{itemize}

Text-based evaluation metrics:
\begin{itemize}
    \item CLIPScore measures the cosine similarity between the CLIP text embedding $c_{clip}$ and CLIP image embedding $v_{clip}$. The metric is computed as $2.5*max(cos(c_{clip}, v_{clip}),0)$. Following the indication of Hessel et al. (2021),  %\cite{DBLP:conf/emnlp/HesselHFBC21},%
    CLIP latent embedding space is computed using a Vision Transformer for image encoding and a Transformer for text encoding.
    
\end{itemize}

\section*{C. Additional results}

Table \ref{tab:results_inpainting} presents the results of the comparison between Stable Diffusion and Blended Latent Diffusion for editing the masked regions detected by DM-Align. According to all image-based and text-based metrics, Stable Diffusion confers more robust editing capabilities than Blended Latent Diffusion and it is therefore used to implement DM-Align. 

The comparison between DM-Align and Prompt-to-Prompt is presented in Table \ref{tab:ptp_clip}. Since Prompt-to-Prompt cannot edit real images but only images generated using the source text instruction, the comparison is limited to CLIPScore. Based on the provided results, the images edited by DM-Align better align with the target text instructions than those edited by Prompt-to-Prompt. The discrepancy in CLIPScore between DM-Align and Prompt-to-Prompt may also be attributed to the fact that the initial images generated by Prompt-to-Prompt do not always reflect the entire source text instructions (images 1 and 4 in Figure \ref{fig:examples_ptp}). 
Regarding the similarity with the input images generated using the source caption, Prompt-to-Prompt encounters difficulties in preserving the background, as one can see in images 6, 8, and 9 of Figure \ref{fig:examples_ptp}.

\end{document}